\documentclass[sigconf]{acmart}

\usepackage{multirow}
\usepackage{fontawesome}
\copyrightyear{2024}
\acmYear{2024}
\setcopyright{rightsretained}
\acmConference[SIGGRAPH Conference Papers '24]{Special Interest Group on Computer Graphics and Interactive Techniques Conference Conference Papers '24}{July 27-August 1, 2024}{Denver, CO, USA}
\acmBooktitle{Special Interest Group on Computer Graphics and Interactive Techniques Conference Conference Papers '24 (SIGGRAPH Conference Papers '24), July 27-August 1, 2024, Denver, CO, USA}
\acmDOI{10.1145/3641519.3657427}
\acmISBN{979-8-4007-0525-0/24/07}

\begin{document}
\title{A Construct-Optimize Approach to Sparse View Synthesis without Camera Pose}

\author{Kaiwen Jiang} 
\email{kevinjiangedu@gmail.com}
\affiliation{%
 \institution{University of California, San Diego}
 \country{United States of America}
}

\author{Yang Fu}
\email{yangfuwork@gmail.com}
\affiliation{%
 \institution{University of California, San Diego}
 \country{United States of America}
 }

\author{Mukund Varma T}
\email{mukundvarmat@gmail.com}
\affiliation{%
 \institution{University of California, San Diego}
 \country{United States of America}
}

\author{Yash Belhe}
\email{yashbelhe2008@gmail.com}
\affiliation{%
 \institution{University of California, San Diego}
 \country{United States of America}
 }

\author{Xiaolong Wang}
\email{xiw012@ucsd.edu}
\affiliation{%
 \institution{University of California, San Diego}
 \country{United States of America}
}

\author{Hao Su}
\email{haosu@ucsd.edu}
\affiliation{%
 \institution{University of California, San Diego}
 \country{United States of America}
}
 
\author{Ravi Ramamoorthi}
\email{ravir@cs.ucsd.edu}
\affiliation{%
 \institution{University of California, San Diego}
 \country{United States of America}
}

\renewcommand{\shortauthors}{Jiang et al.}


\begin{teaserfigure}
  \centering
  \includegraphics[width=\linewidth]{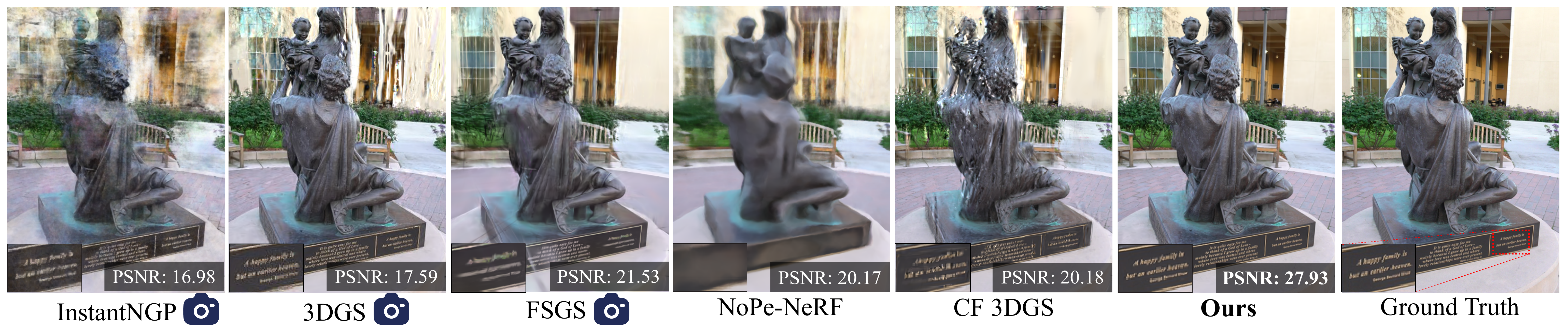}
  \caption{
  {We introduce a sparse view synthesis method, which does not rely on off-the-shelf estimated camera poses. }
  Given the ``Family'' scene in the Tanks \& Temples dataset, we use $6$ out of $200$ frames as training views {and others for testing.}
  {Compared with other pose-free methods, including COLMAP-Free 3DGS \cite{colmap-free} (CF 3DGS) and NoPe-NeRF \cite{nopenerf}, we achieve significant improvements in novel view synthesis both qualitatively and quantitatively. 
  Besides, we also outperform methods which rely on off-the-shelf estimated camera poses, including Instant-NGP \cite{instant-ngp}, Gaussian Splatting \cite{3DGS} (3DGS), and FSGS \cite{fsgs}.%
  }
  Methods marked with a camera rely on off-the-shelf estimated camera poses throughout the paper. The inscription under the statue is emphasized to compare high-frequency details.
  Image credits by \citet{tanks_and_temples}.
  }
  \label{fig:teaser}
\end{teaserfigure}

\begin{abstract}
Novel view synthesis from a sparse set of input images is a challenging problem of great practical interest, especially when camera poses are absent or inaccurate. 
Direct optimization of camera poses {and usage of} estimated depths in neural radiance field algorithms usually do not produce good results because of the coupling between poses and depths, and inaccuracies in monocular depth estimation. 
In this paper, we leverage the recent 3D Gaussian splatting method to develop a novel construct-and-optimize method for sparse view synthesis without camera poses. Specifically, we \emph{construct} a solution {progressively} by using monocular depth and projecting pixels back into the 3D world.  {During construction, }we \emph{optimize} the solution by detecting 2D correspondences between training views and the corresponding rendered images. We develop a unified differentiable pipeline for camera registration and adjustment of {both camera poses and }depths, followed by back-projection. 
We also introduce a novel notion of an expected surface in Gaussian splatting, which is critical to our optimization. 
These steps enable a coarse solution, which can then be low-pass filtered and refined using standard optimization methods. 
We demonstrate results on the Tanks and Temples and Static Hikes datasets with as few as three widely-spaced views, showing significantly better quality than competing methods, including those with approximate camera pose information.  Moreover, our results improve with more views and outperform previous InstantNGP and Gaussian Splatting algorithms even when using half the dataset. \emph{Project page: https://raymondjiangkw.github.io/cogs.github.io/}
\end{abstract}



\maketitle

\section{Introduction}
{Neural Radiance Field (NeRF) and its several variants~\cite{nerf2020, 3DGS, instant-ngp, zip-nerf} have excelled in novel view synthesis of 3D scenes. 
However, these methods require densely captured views with accurately labeled camera poses, which is often not feasible in practical scenarios. 
Often, camera poses are obtained from Structure-from-Motion (SfM) methods like COLMAP~\cite{sfm1, sfm2} as a pre-processing step to NeRF, which is brittle and fails when given sparse views. 
{Even in the cases where COLMAP can successfully register sparse scenes, as shown in Fig.~\ref{fig:teaser}, %
{sparse view synthesis is {challenging and} ill-posed from ambiguity in the 3D scene due to under-sampling. }
This limitation raises a critical question: Is it possible to perform novel view synthesis {from sparse input captures (as little as 3-6 images) with unknown camera poses}?}}

{The recent introduction of 3D Gaussian Splatting, denoted as 3DGS~\cite{3DGS}, also struggles to deal with sparse view synthesis due to too sparse initialization from SfM. }
{However, the explicit representation, i.e., 3D Gaussians, of 3DGS {provides new opportunities to solve that critical question.}}
Different from fitting a solution for sparse view synthesis in NeRFs, we desire to \emph{construct} a solution based on a dense prior, i.e., estimated monocular depth; however {\em optimization} is still essential, and we therefore call our approach a construct-and-optimize method.

\begin{figure}[t!]
    \centering
    \includegraphics[width=\linewidth]{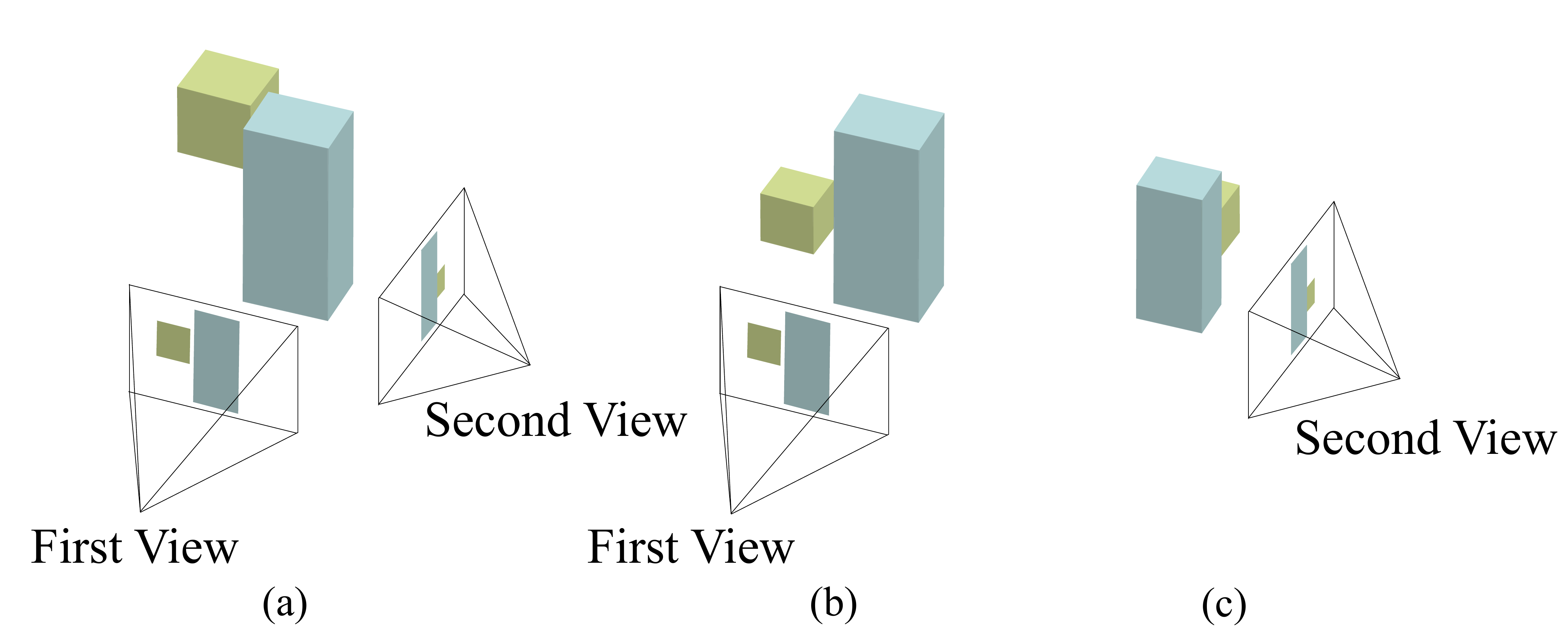}
    \caption{Example of ambiguity given partial views. Given the scene in (a), there could be different possibilities of scene layouts as shown in (b) and (c), if only the first view or second view is observed. 
    (b) or (c) could be the estimated depth. 
    This ambiguity results in unavoidable error in monocular depth estimation, which necessitates the alignment between camera poses and estimated depths.
    }
    \label{fig:ambiguity}
\end{figure}

\begin{figure}[t!]
    \centering
    \includegraphics[width=\linewidth]{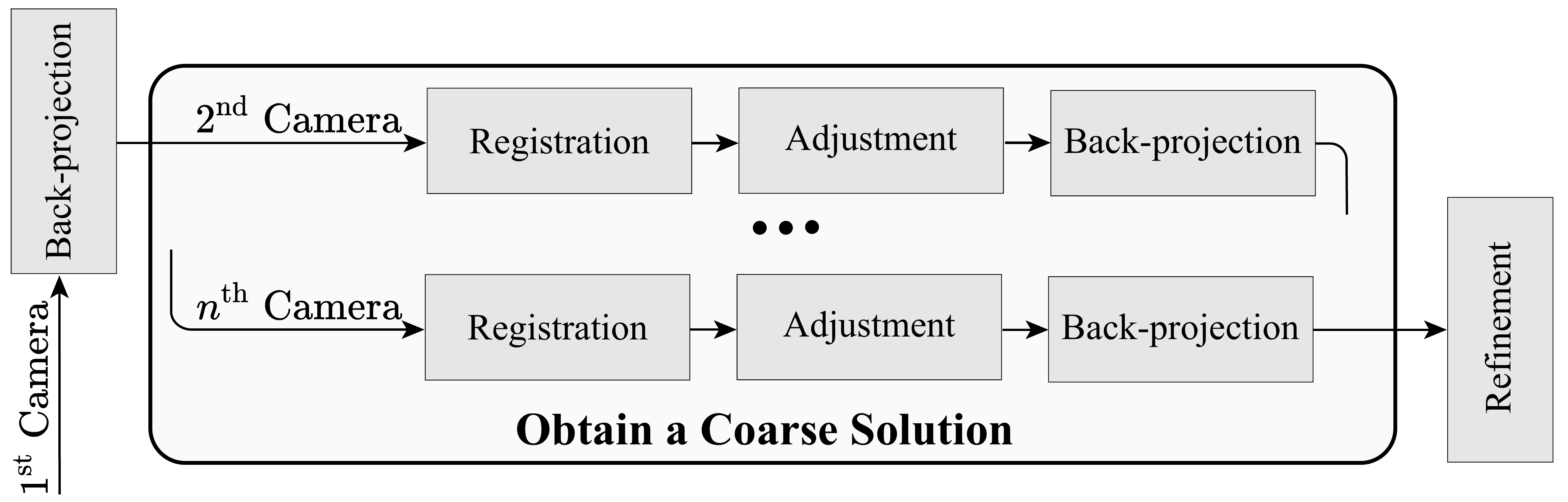}
    \caption{Overview of our method for sparse view synthesis. We first back-project the first view and sequentially {register}, {adjust} and {back-project} the remaining views in sequence to obtain a coarse solution. This coarse solution is then refined by standard optimization to reproduce fine details.}
    \label{fig:steps-overview}
\end{figure}

{A naive way to {construct a solution} is by first estimating camera poses and then back-projecting pixels into the scene based on their estimated depths. }
{However, there is a problem in handling camera poses and depth estimation independently -- in actual 3D captures, both quantities are tightly coupled and depend upon one another \cite{consistent_video_depth} as shown in Fig.~\ref{fig:ambiguity}. }
{Unfortunately, }in the case of sparse views, {monocular depth estimation algorithms do not take camera pose information into account.
Additionally, camera pose estimation algorithms do not leverage and align monocular depth. As a result, back-projection for the same scene across multiple views may be inconsistent. }
We {therefore involve optimization in the construction} {to solve these issues.} 
{Our pipeline shown in Fig.~\ref{fig:steps-overview} is progressive, i.e. it builds the scene continuously.}
{For the next unregistered view, we first estimate its camera pose in a \textit{registration} stage. Afterwards, we adjust the previous registered camera poses and align monocular depths, which we call \textit{adjustment}.
{{At last}, pixels of {the next view} are back-projected into world space as {3D Gaussians}.} Therefore, the camera poses are not needed to be known in advance.
{Finally, we {reach a coarse solution, which is then refined using standard optimization \cite{3DGS} to reproduce details faithfully. Before that, we also apply a low-pass filtering to avoid high-frequency artifacts as shown in Fig.~\ref{fig:ablation-study} (a).}
}}

\begin{figure*}[t]
    \centering
    \includegraphics[width=\linewidth]{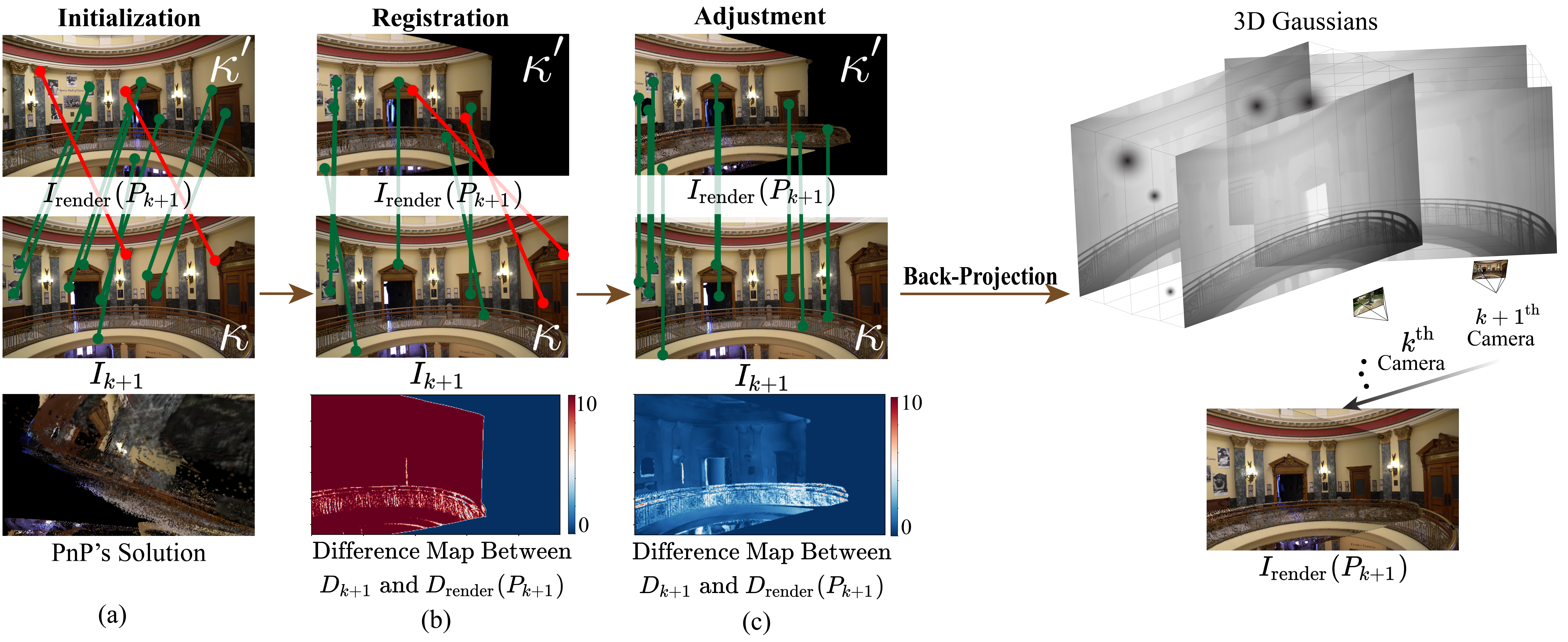}
    \caption{
    We assume the first $k$ views have already been registered, and {illustrate the registration, adjustment and back-projection of the $k+1$\textsuperscript{th} view}. 
    {
    \textbf{(a)} We first initialize the camera pose of the $k+1$\textsuperscript{th} view, denoted as $P_{k+1}$, as the $k$\textsuperscript{th} view's camera pose. 2D correspondences are detected between ground-truth image $I_{k+1}$ and the rendered result $I_\text{render}(P_{k+1})$ at $P_{k+1}$. 
    Correspondence points on $I_\text{render}(P_{k+1})$ are denoted as $\kappa'$, while those on $I_{k+1}$ are denoted as $\kappa$. 
    Green points denote correct correspondences, while red points denote wrong correspondences. 
    We can use perspective-n-points (PnP) to solve the camera pose but it results in an erroneous solution. 
    \textbf{(b)} We then apply our optimization pipeline (Sec.~\ref{methodology-optimization}) to estimate the camera pose for registration. For now, the monocular depth $D_{k+1}$ of the $k+1$\textsuperscript{th} view deviates significantly from the rendered depth $D_\text{render}(P_{k+1})$ at $P_{k+1}$. 
    \textbf{(c)} Afterwards, we apply our optimization pipeline (Sec.~\ref{methodology-optimization}) to adjust all previous registered camera poses and monocular depths along with $P_{k+1}$ and $D_{k+1}$. It can be seen that $I_\text{render}(P_{k+1})$ and $D_{k+1}$ are much close to $I_{k+1}$ and $D_\text{render}(P_{k+1})$. 
    Finally, we back-project pixels in the $k+1$\textsuperscript{th} view into world space as {3D Gaussians} based on $D_{k+1}$. %
    }
    Images credit by \citet{tanks_and_temples}.
    }
    \label{fig:pipeline}
\end{figure*}

{To harmonize the monocular depths with the camera poses, we}
unify \emph{registration} and \emph{adjustment} in a differentiable pipeline for optimization{, whose objective is to {reproduce training views}}. {However, for sparse views with non-trivial camera movements, commonly used pixel-wise supervision {does not} lead to effective optimization{, since it only considers short-range information}. 
}
{To take long-range information into account, we instead detect 2D correspondences \cite{quadtree, loftr} between training views and their corresponding rendered views, and aim to match them in screen-space coordinates as shown in Fig.~\ref{fig:pipeline} (a), (b). }
{To optimize over the detected correspondences, {we have to render screen-space coordinates {of expected surface points}, which requires }a definition of a{n expected} surface {in Gaussian splatting}. Existing attempts at estimating the {expected} surface (e.g., \cite{depth-gs, fsgs, splatam}) do not fully respect the shape of Gaussian kernels, leading to ineffective optimization for our problem. We are therefore motivated to develop a more accurate {rendering of the expected surface} for 3DGS.}

Through extensive experiments and comparisons, we show that our method achieves state-of-the-art results when dealing with challenging cases where only {a few} views {($<5\%$ of total views)} are provided and there is non-trivial camera movement between any pair of views{, as shown in Fig.~\ref{fig:teaser},~\ref{fig:qualitative}}. The performance of our method also improves as the number of views increases {as shown in Fig.~\ref{fig:number-of-views}}. 

In summary, the contributions of our work include:
\begin{itemize}
    \item We {propose a unified differentiable pipeline, which leverages correspondences as supervision, for %
    {sparse view synthesis without camera poses in Sec.~\ref{methodology-depth-back-projection},~\ref{methodology-optimization}}.
    \item {We propose a} differentiable approximation of {the expected} surface in Gaussian splatting for {{effective} correspondence supervision}{ in Sec.~\ref{methodology-differentiable-approximate-surface-rendering}}.}
\end{itemize}

\section{Related Work}
~

\paragraph{Sparse view synthesis.}{The computer vision and graphics community has studied novel view synthesis for decades \cite{related_classic_work_1, related_classic_work_2, related_classic_work_3, related_classic_work_6, related_classic_work_7}. {A number of subsequent advances were made in sparse view synthesis \cite{related_classic_work_5, llff, instant-3d-photography}.} {We build on the recent development of neural radiance fields for view synthesis \cite{nerf, nerf2020}}. }

{Sparse view synthesis in NeRFs} typically assumes camera poses to be known to simplify the problem. 
Some works (e.g., \cite{info-nerf, reg-nerf, sparse-nerf, DSNeRF, MonoSDF, pixel-nerf}) try to fit an efficient representation (e.g., MLPs \cite{nerf2020}, hash tables \cite{instant-ngp}, etc.) with prior knowledge, such as depth or heuristic constraints, to reduce the ambiguity. However, the reconstructed continuous signal still unavoidably results in blurry or noisy images for novel views. %

In this paper, we show that {constructing a solution} for {sparse view synthesis} {with optimization} can be easier, in contrast to fitting the complex signal from sparse observations. 
However, the camera poses {and the reconstructed scene should be aligned. }%
{In SfM%
, this goal is achieved by \emph{bundle adjustment} \cite{bundle_adjustment}. However, our representation for the scene as 3D Gaussian is different from points.
Realistic rendering and the higher degrees of freedom afforded by 3D Gaussians enable correspondence detection between the current reconstructed scene and {training views} %
for alignment.
This facilitates effective and stable optimization based on differentiable rendering.
Therefore, we call this optimization process \emph{adjustment} to distinguish it from conventional bundle adjustment.
Furthermore, \emph{registration} of camera poses can also be unified in the same optimization framework. The camera poses are therefore not needed to be known in advance. }
\paragraph{Optimizing camera poses in NeRFs.}Accurate camera poses are vital for realistic {view synthesis} in NeRFs. 
Given initially estimated camera poses, which usually come from SfM \cite{sfm1, sfm2} or sensors, some methods (e.g., \cite{camp, barf, robust_refinement, scanerf}) refine them during the optimization for better {view synthesis}.

When camera poses are not given, several works \cite{nerfmm, nopenerf, localrf, colmap-free, barf} require a dense capture and gradually register frames by {pixel-wise} supervision and/or additional priors, such as depth \cite{midas1, midas2} or optical-flow \cite{raft}. It is noteworthy that the camera pose of the next unregistered frame is initialized as the camera pose of the last registered frame. 
However, when the views are sparse and there is non-trivial camera movement between any pair of captured frames, SfM sometimes fails to produce accurate results and none of the previous methods can deal with this scenario well. GNeRF \cite{gnerf} tries to tackle the general camera pose querying problem by a generative prior, which is still limited to individual objects rather than scenes. 

As discussed in Xing et al.~\shortcite{rgb-xy}, {pixel-wise} supervision yields gradients that only consider short-range information. In the case of sparse views, there could be non-trivial camera movements that makes it desirable to have gradients that consider long-range information. 
{Therefore, in this work, we {leverage 2D} correspondences between the reconstructed scene and {training views} for effective optimization of camera poses and the reconstructed scene.}
\paragraph{Surface rendering in Gaussian splatting.}In 3DGS, recent progress (e.g., \cite{depth-gs, colmap-free, fsgs, sparse_gs, splatam, gs-slam}) has found that an approximate surface is useful for {view synthesis}.
However, their approximation unavoidably assumes that the surface is a constant point inside each Gaussian kernel, which is sub-optimal in our case.
Therefore, we propose an approximate anisotropic surface rendering scheme that is more accurate than prior works and results in effective optimization for our task.
A variant of our approximate surface rendering is introduced in \cite{pulsar}, but it is isotropic and does not deal with volume rendering. 
SuGaR \cite{sugar} proposes an ideal depth to regularize the current rendered depth.
In contrast, we better approximate the expected surface for downstream optimization.

\section{Method}
\label{methodology}
{In Sec.~\ref{methodology-depth-back-projection}, we first present an overview of our algorithm {for sparse view synthesis}. 
In Sec.~\ref{methodology-optimization}, we introduce our differentiable pipeline for registration and adjustment.
Next, we introduce a more accurate approximation of surface rendering for 3D Gaussians in Sec.~\ref{methodology-differentiable-approximate-surface-rendering}, which allows us to leverage correspondences as an effective supervision in the differentiable pipeline. 
After these steps, we have a coarse solution, which we further refine in %
Sec.~\ref{sec:refinement}. %
We present an outline of our pipeline in Fig.~\ref{fig:steps-overview}, and focus on registering, adjusting and back-projecting the $k+1$\textsuperscript{th} view in Fig.~\ref{fig:pipeline}.
}

\subsection{Algorithm Overview}
\label{methodology-depth-back-projection}{
Assuming we have an ordered set of consecutively captured $n$ RGB images $\mathcal{I}=\{I_1, I_2, ..., I_n\}$ and their corresponding estimated monocular depths $\mathcal{D}=\{D_1, D_2, ..., D_n\}$, we are interested in novel view synthesis {without camera poses}, using 3D Gaussians as our representation. 
{As in \cite{nopenerf}, we assume the intrinsic matrix $K$ of the camera is given, and denote the unknown extrinsic matrices for each view as $\mathcal{P}=\{P_1, P_2, ..., P_n\}$.}

As shown in Fig.~\ref{fig:steps-overview}}, we start with the first view $I_1$ and set its extrinsic matrix $P_1$ to the identity matrix.
Next, we back-project each of its pixels into world space as {3D Gaussians}, such that the rendered image and depth match $I_1$ and $D_1$ respectively. 

Specifically, given the camera pose and depth for the frame, we can construct a particular fully opaque splat for each pixel in our approximate surface rendering scheme (please find details in Sec. 1.1 of the supplementary). 

{We then assume }the first $k$ frames have already been registered {and consider the next unregistered view $I_{k+1}$}. 
Its extrinsic matrix $P_{k+1}$ is first initialized as $P_k$ as shown in Fig.~\ref{fig:pipeline} (a). We optimize $P_{k+1}$ based on the previous reconstruction to register the new view as shown in Fig.~\ref{fig:pipeline} (b).
During adjustment, we optimize all previous extrinsic matrices $\{P_1, P_2, ..., P_k\}$ and monocular depths $\{D_1, D_2, ..., D_k\}$ along with $P_{k+1}$ and $D_{k+1}$ as shown in Fig.~\ref{fig:pipeline} (c).
After that, pixels in $I_{k+1}$ are back-projected based on the aligned depth $D_{k+1}$. 
After processing all $n$ views, we reach a coarse solution for sparse view synthesis.

\subsection{Optimization Framework}
\label{methodology-optimization}
{
{We achieve \emph{registration} and \emph{adjustment} through an optimization framework.}
The camera pose is optimized in both \emph{registration} and \emph{adjustment}, but the alignment of depth is achieved through the \emph{adjustment} only.
Our optimization aims to reproduce training views, i.e., for each view, the rendered image should match the ground-truth image. 
Pure pixel-wise supervision does not lead to effective optimization for non-trivial camera movement between consecutive views, since it only considers short-range information. 
Inspired by but different from optimal transport \cite{rgb-xy}, we leverage correspondences, which bootstrap the method, to consider long-range information for optimization. %

Assume the view of interest is $I$, and the rendered image given its current extrinsic matrix $P$, is $I_\text{render}(P)$. 
We can leverage off-{th}e-shelf detectors \cite{quadtree, loftr} to detect the 2D correspondences between $I$ and $I_\text{render}(P)$ per optimization step. The 2D screen-space points on $I$ are $\mathbf{\kappa}=\{\mathbf{\kappa}^{(1)}, \mathbf{\kappa}^{(2)}, ..., \mathbf{\kappa}^{(M)}\}$, where $M$ denotes the number of points. The 2D screen-space points on $I_\text{render}(P)$ are $\mathbf{\kappa}'=\{\mathbf{\kappa}^{'(1)}, \mathbf{\kappa}^{'(2)}, ..., \mathbf{\kappa}^{'(M)}\}$. The optimization goal is then to match $\mathbf{\kappa}$ with $\mathbf{\kappa}'$, which is visualized in Fig.~\ref{fig:pipeline} (a) and (b). 
{
For registration only, we can use perspective-n-points (PnP) \cite{epnp} to solve camera parameters.
However, it is sensitive to mismatches as shown in Fig.~\ref{fig:pipeline} (a), and it is hard to balance the number of matches with the threshold of the RANSAC algorithm.
On the other hand, we find that our optimization framework is robust and achieves more accurate registration.
Therefore, registration and adjustment are unified under the same optimization framework.}

In order to back-propagate gradients from the matching between $\mathbf{\kappa}$ and $\mathbf{\kappa}'$ to the 3D Gaussians that form the surface, we use a differentiable approximate surface renderer, elaborated in Sec.~\ref{methodology-differentiable-approximate-surface-rendering}, to render screen-space coordinates at $\mathbf{\kappa}^{'(i)}, i=1,2,...,M$ as $q(c^{'})$.
The resulting loss is
\begin{equation}
    \mathcal{L}_\text{corr} = \sum_{i=1}^M || q(\mathbf{\kappa}^{'(i)}) - \mathbf{\kappa}^{(i)} ||_1.
\end{equation}
Importantly, $q(\mathbf{\kappa}^{'(i)})$ equals $\mathbf{\kappa}^{'(i)}${ numerically, but it forms a graph for back-propagating gradients to the underlying representation}. 

We also find that when $P$ is close to ground-truth, short-range information provided by pixel-wise supervision can help stabilize the optimization. 
This loss is given by
\begin{equation}
    \mathcal{L}_\text{rgb} = || I - I_\text{render}(P) ||_1.
\end{equation}

Finally, in the \emph{adjustment}, we adjust the monocular depth $D$ of the current view {during the adjustment phase}, for later use during back-projection. 
To {align} the estimated monocular depths effectively, we would like to use the scale-consistency assumption \cite{midas2,MariGold} but it does not always hold true.
To relax the scale-consistency assumption, instead of learning an affine transformation \textit{per view}, we learn a separate affine transformation \textit{per primitive} \cite{fcclip}. 
By denoting the rendered depth given the current extrinsic matrix $P$ as $D_\text{render}(P)$, we match $D$ to the current rendered depth $D_\text{render}(P)$ for all correspondences. 
Specifically, the rendered depth at $\mathbf{\kappa}^{'}$ is denoted as $b(\mathbf{\kappa}^{'})$ and the monocular depth at $\mathbf{\kappa}$ is denoted as $d(\mathbf{\kappa}^{})$.
}
The loss term is defined as
\begin{equation}
    \mathcal{L}_\text{depth} = \sum_{i=1}^M || \text{sg}[b(\mathbf{\kappa}^{'(i)})] - d(\mathbf{\kappa}^{(i)}) ||_1. 
\end{equation}
Since we only want to adjust the monocular depth $D$ to make it match the rendered depth $D_\text{render}(P)$, we stop the backward propagation of gradients ($\text{sg}[\cdot]$) through $b$.

In summary, the optimization objective is defined as
\begin{equation}
\label{eqn:optimization-objective}
    \mathcal{L} = \lambda_1 \mathcal{L}_\text{corr} + \lambda_2 \mathcal{L}_\text{rgb} + \lambda_3 \mathcal{L}_\text{depth}, 
\end{equation}
where $\lambda_1 = 1000, \lambda_2 = 10, \lambda_3 = 1$. The gradients back-propagate to camera parameters and the reconstructed scene.

\subsection{Differentiable Surface Rendering}
\label{methodology-differentiable-approximate-surface-rendering}
{
Our goal for correspondence matching (see Sec.~\ref{methodology-optimization}) is to propagate long-range gradient information from a 2D screen-space point to its corresponding 3D surface point.
In essence, our goal is to map perturbations of the 2D screen-space point to corresponding perturbations of the 3D surface point.
However, this raises an important question -- where is the 3D surface point? %

\begin{figure}[t]
    \centering
    \includegraphics[width=\linewidth]{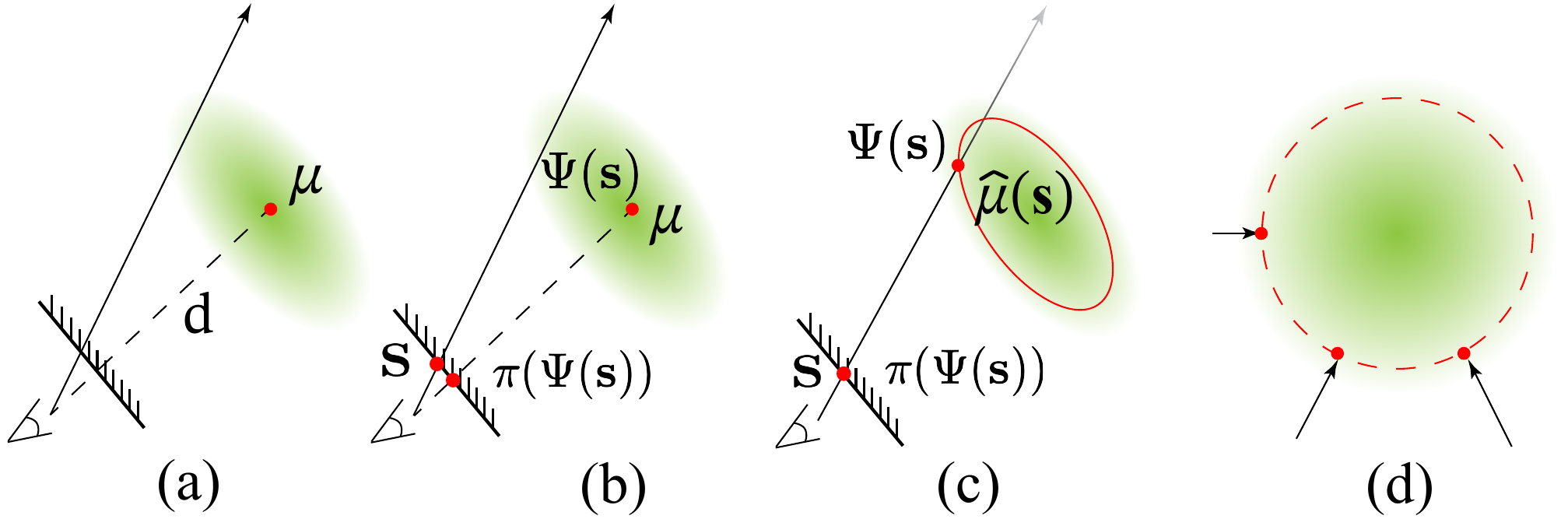}
    \caption{Illustration of surface rendering in Gaussian splatting. {Assume the ray is shot from screen-space coordinates $\mathbf{s}$ and $\Psi(\mathbf{s})$ denotes the rendered surface point. $\pi(\cdot)$ denotes projecting 3D points into screen space.} (a) Depth rendering of previous methods. The depth $d$ of a Gaussian kernel is defined as the $z$-axis coordinate for the transformed center $\boldsymbol{\mu}$ in the camera space. (b) Extending (a) to render the exact 3D surface point. The surface {point} of the Gaussian kernel is defined as the center $\boldsymbol{\mu}$. It could result in a mismatch between $\mathbf{s}$ and $\pi(\Psi(\mathbf{s}))$. 
    (c) Approximate surface rendering of our method. The surface point $\widehat{\boldsymbol{\mu}}(\mathbf{s})${ of the Gaussian kernel} is defined as the intersection point between the ray and an ellipsoid shell.{ Therefore, our method guarantees a match between $\mathbf{s}$ and $\pi(\Psi(\mathbf{s}))$.} 
    (d) Surface rendering of our method when considering all the rays passing through the center of a spherical Gaussian kernel. The {expected }surface points form a shell. %
    }
    \label{fig:surface-rendering}
\end{figure}

Since the 3D Gaussian representation is volumetric, there are no explicit surfaces; instead, previous works \cite{splatam, depth-gs, colmap-free, gs-slam} compute the depth of \textit{expected} 3D surface point $D(\mathbf{s})$ corresponding to the 2D screen-space point $\mathbf{s}$ as:
\begin{equation}
\label{equation-native-depth}
\begin{aligned}
    D(\mathbf{s}) &= \sum_{i} d_i \alpha_i(\mathbf{s}) \prod_{j=1}^{i-1} (1 - \alpha_j(\mathbf{s})),
\end{aligned}
\end{equation}
where $d_i\in\mathbb{R}$ denotes the $z$-axis coordinate for the transformed Gaussian centers in the camera space, and $\alpha_i\in\mathbb{R}$ and $\alpha_j\in\mathbb{R}$ denote the calculated alpha-blending coefficient of the $i$\textsuperscript{th} and $j$\textsuperscript{th} Gaussian kernel. This is shown as Fig.~\ref{fig:surface-rendering} (a).

Its extension to corresponding \textit{expected} 3D surface point $\Psi(\mathbf{s})$ at $\mathbf{s}$ is given by
\begin{equation}
\begin{aligned}
    \Psi(\mathbf{s}) &= \sum_{i} {\boldsymbol{\mu}}_i \alpha_i(\mathbf{s}) \prod_{j=1}^{i-1} (1 - \alpha_j(\mathbf{s})), 
\end{aligned}
\end{equation}
where ${\boldsymbol{\mu}}_i\in\mathbb{R}^3$ denotes the center of the $i^\mathrm{th}$ Gaussian kernel.

Unfortunately, their rendering model for $\Psi(\mathbf{s})$ is not consistent with $\mathbf{s}$, i.e., $\Psi(\mathbf{s})$ may not be projected onto $s$, which is essential to our optimization, see Fig.~\ref{fig:surface-rendering} (b).

We would like to fix this rendering model, without breaking any of the original assumptions~\cite{3DGS, ewa, surface_splatting} in order to have efficient rendering.
Our solution is to replace $\boldsymbol{\mu}_i$ with a better approximation $\widehat{\boldsymbol{\mu}}_i(\mathbf{s})$ for {the expected} surface point.
Different from the earlier rendering model, $\widehat{\boldsymbol{\mu}}_i(\mathbf{s})$ is now dependent on $\mathbf{s}$, as illustrated in Fig.~\ref{fig:surface-rendering} (c).

We expect the relative position 
$\delta$ between $\widehat{\boldsymbol{\mu}}_i(\mathbf{s})$ and $\boldsymbol{\mu}_i$ to be translation- and rotation-invariant, as shown in Fig.~\ref{fig:surface-invariant} (a). We can therefore consider $\widehat{\boldsymbol{\mu}}_i(\mathbf{s})$ in the canonical form, i.e., the Gaussian kernel is placed canonically at the origin. 
For a single isotropic 3D Gaussian, we can compute the expected surface point using its free flight distance probability density function.
The locus of expected surface points for all rays passing through the center of the 3D Gaussian form a shell, as shown in Fig.~\ref{fig:surface-rendering} (d).
We use this shell to approximate the surface of the 3D Gaussian.
Similarly, for an anisotropic 3D Gaussian, we use an ellipsoidal shell to approximate the surface.
Here, the semi-axes can be computed by an integral which only depends on the anisotropic Gaussian's scale parameters.
For efficient gradient calculation, we approximate this integral with a linear function. 
Our rendering guarantees the ``consistency'' property, {i.e., $\Psi(\mathbf{s})$ is projected to $\mathbf{s}$,} as shown in Fig.~\ref{fig:surface-rendering} (c).

\begin{figure}[t]
    \centering
    \includegraphics[width=\linewidth]{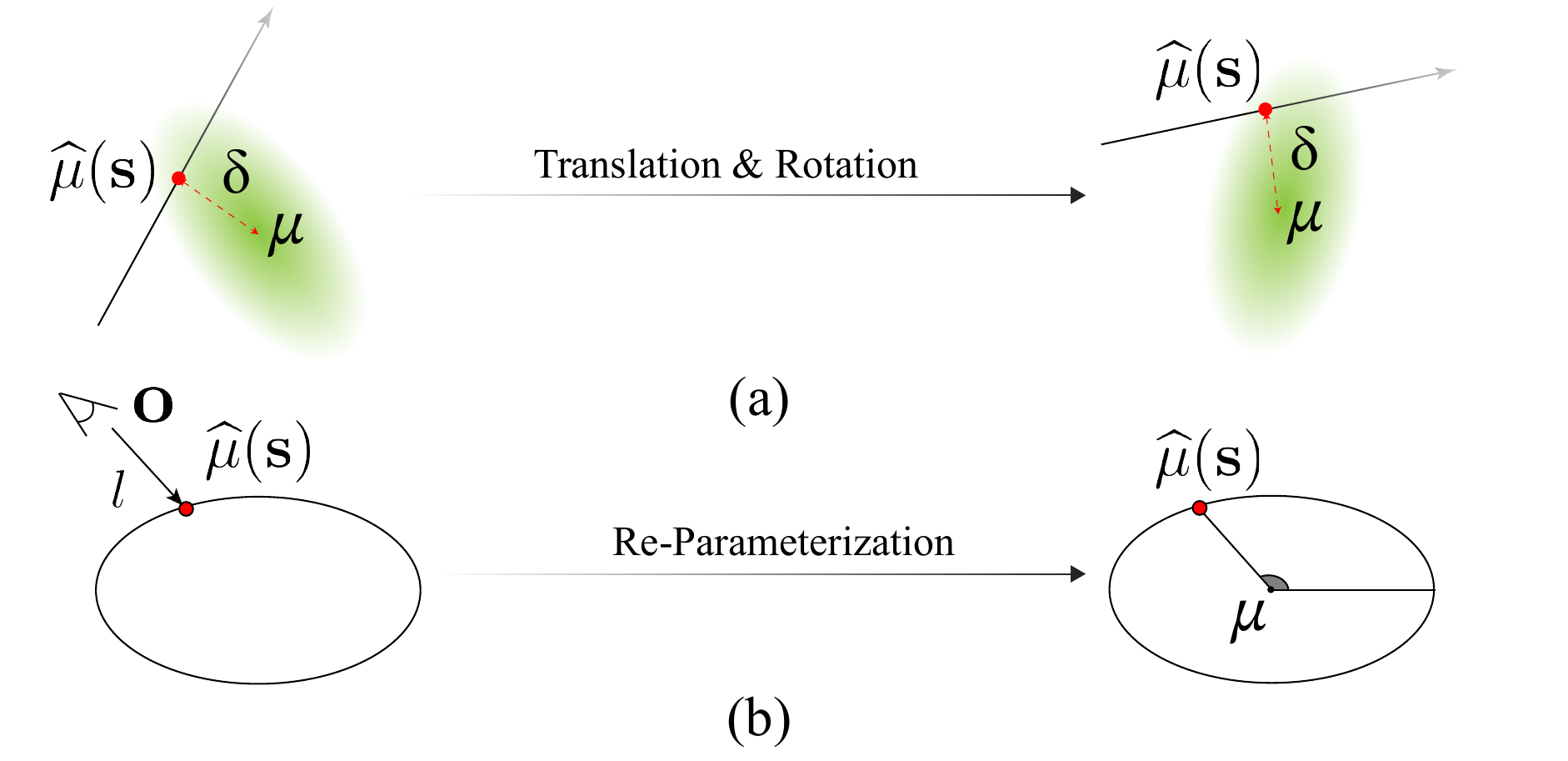}
    \caption{(a) Illustration of the invariance of relative position $\delta$ between the surface point $\widehat{\boldsymbol{\mu}}(\mathbf{s})$, and the center of the Gaussian kernel $\boldsymbol{\mu}$. $\delta$ is expected to be translation- and rotation-invariant.
    {(b) Illustration of re-parameterizing surface point $\widehat{\boldsymbol{\mu}}(\mathbf{s})$ from an intersected point into a point defined on an ellipsoid shell.}
    }
    \label{fig:surface-invariant}
\end{figure}

In summary, $\widehat{\boldsymbol{\mu}}_i(\mathbf{s})$ is reduced to a ray-intersection test between the current ray and a defined ellipsoid shell of the current Gaussian kernel, illustrated in Fig.~\ref{fig:surface-rendering} (c). 
Specifically, assume the origin and corresponding direction of the current ray are $\mathbf{o}$ and $\mathbf{d}(\mathbf{s})$, and the intersection distance is $l$, so that $\widehat{\boldsymbol{\mu}}_i(\mathbf{s}) = \mathbf{o} + l\mathbf{d}(\mathbf{s})$.
And we skip Gaussian kernels whose approximate surfaces do not intersect with the current ray, denoted as $\widehat{\boldsymbol{\mu}}_i(\mathbf{s})=\varnothing$. The {expected world space coordinates of the }surface {point at $\mathbf{s}$} are therefore given by
\begin{equation}
\label{equation-surface}
\begin{aligned}
    \Psi(\mathbf{s}) &= \sum_{i, \widehat{\boldsymbol{\mu}}_i(\mathbf{s})\neq~\varnothing} \widehat{\boldsymbol{\mu}}_i(\mathbf{s}) \alpha_i(\mathbf{s}) \prod_{j=1, \widehat{\boldsymbol{\mu}}_j(\mathbf{s})\neq~\varnothing}^{i-1} (1 - \alpha_j(\mathbf{s})).
\end{aligned}
\end{equation}
{By projecting $\widehat{\boldsymbol{\mu}}_i(\mathbf{s})$ into camera space to obtain the $z$-axis coordinate{ as $z_i(\mathbf{s})$} and into screen space to obtain the screen-space coordinates $\pi(\widehat{\boldsymbol{\mu}}_i(\mathbf{s}))$, we are able to render depth $D(\mathbf{s})$ and screen-space coordinates $q(\mathbf{s})$ of expected surface points{, which are given by}}
\begin{equation}
\begin{aligned}
    D(\mathbf{s}) &= \sum_{i, \widehat{\boldsymbol{\mu}}_i(\mathbf{s})\neq~\varnothing} z_i(\mathbf{s}) \alpha_i(\mathbf{s}) \prod_{j=1, \widehat{\boldsymbol{\mu}}_j(\mathbf{s})\neq~\varnothing}^{i-1} (1 - \alpha_j(\mathbf{s})) \\
    q(\mathbf{s}) &= \sum_{i, \widehat{\boldsymbol{\mu}}_i(\mathbf{s})\neq~\varnothing} \pi(\widehat{\boldsymbol{\mu}}_i(\mathbf{s})) \alpha_i(\mathbf{s}) \prod_{j=1, \widehat{\boldsymbol{\mu}}_j(\mathbf{s})\neq~\varnothing}^{i-1} (1 - \alpha_j(\mathbf{s})).
\end{aligned}
\end{equation}

We follow the same framework of forward and backward passes in the rasterizer proposed in \cite{3DGS}, but replace the rendering of color with our defined {expected} surface. 
{However, the backward pass is a bit different due to the involved ray-intersection test. }
Specifically, considering the surface point $\widehat{\boldsymbol{\mu}}_i(\mathbf{s})$ for the $i$\textsuperscript{th} Gaussian kernel, it is defined as $\mathbf{o} + l\mathbf{d}(\mathbf{s})$, in which $l$ is a function of the center, rotation, and scaling of the $i^\mathrm{th}$ Gaussian kernel. 
{This parameterization} could result in gradient cancellation when camera parameters are also optimized, illustrated in Fig.~\ref{fig:gradient-cancellation}. 
Even though the surface point $\widehat{\boldsymbol{\mu}}_i(\mathbf{s})$ is defined through the origin and direction of the current ray, it should be treated as an independent point existing on an ellipsoid shell. 
Therefore, we propose to re-parameterize $\widehat{\boldsymbol{\mu}}_i(\mathbf{s})$ as a function of the center, rotation and scaling of the ellipsoid shell solely{ as shown in Fig.~\ref{fig:surface-invariant} (b)}. 
The gradients are then back-propagated to these properties of the ellipsoid shell, and finally to the camera parameters and the Gaussian kernel.
}

\subsection{Refinement and Implementation Details}
\label{sec:refinement}
{
After reaching a coarse solution using the algorithm above, we refine the solution using standard optimization techniques \cite{3DGS} and optimize the camera poses in the optimization. 
Before that, we first remove error-prone high-frequency reconstructions  by applying a low-pass filter, see Fig.~\ref{fig:ablation-study} (a).
To achieve the same effect as a low-pass filter by penalizing dense clusters of Gaussians and promoting more uniformly distributed Gaussians, for each detected object in each view, we only retain $10\%$ of the total back-projected Gaussians, by a farthest point sampling algorithm \cite{pytorch3d} based on the center of the Gaussians. 
}

\begin{figure}[t]
    \centering
    \includegraphics[width=\linewidth]{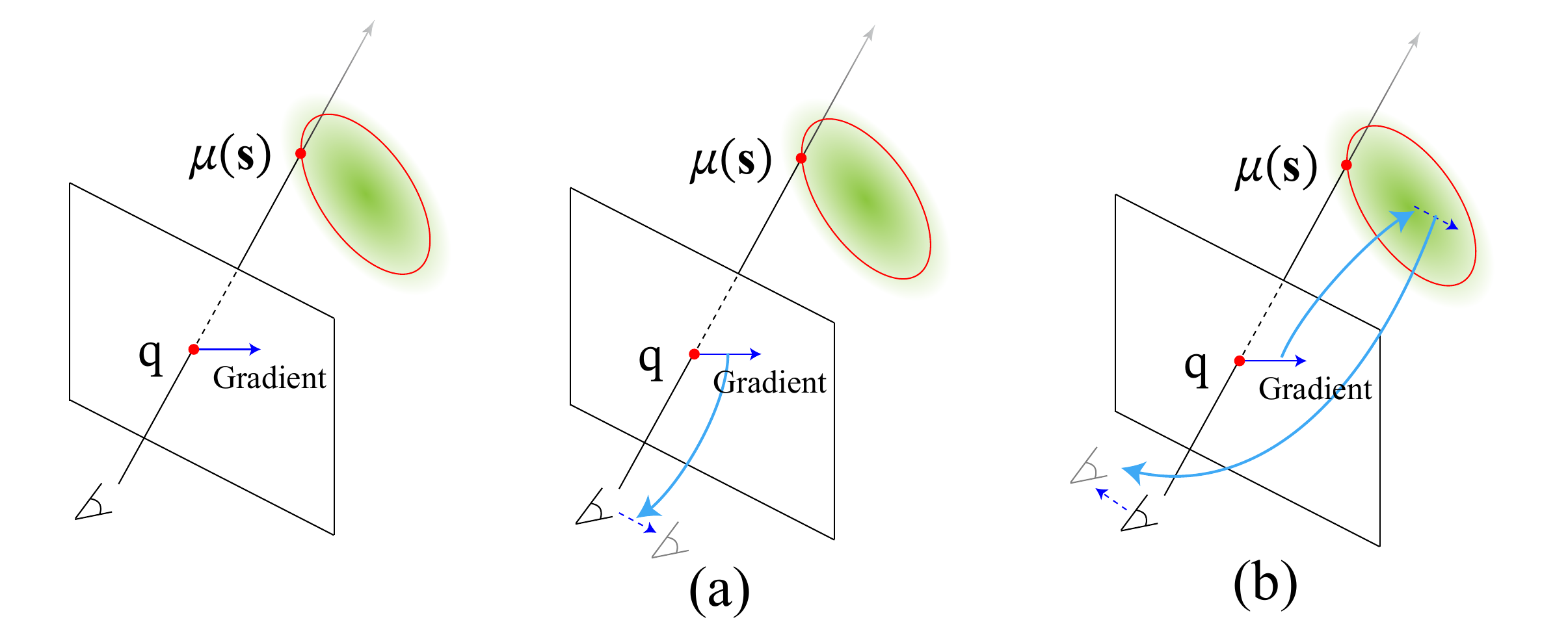}
    \caption{Illustration of gradient cancellation due to the intersection test. Given the surface point $\boldsymbol{\mu}(\mathbf{s})$ and its projected screen-space coordinates $q$, assume the gradients move $q$ to the right. 
    (a) In one way, since $\boldsymbol{\mu}(\mathbf{s})$ is defined through the ray origin and direction, the gradients are back-propagated to the ray origin to move the viewing position right. 
    (b) In another way, since $\boldsymbol{\mu}(\mathbf{s})$ is defined through the center of the Gaussian kernel, the gradients are back-propagated to the center to move it to the right. In turn, due to the transformation from world space to camera space, the gradients are back-propagated to the viewing position to move it left. 
    Therefore, these two gradients cancel each other and sum to zero occasionally.
    }
    \label{fig:gradient-cancellation}
\end{figure}

The optimization setup is almost the same as Kerbl et al.~\shortcite{3DGS}.
{On an NVIDIA RTX 3080 GPU, the time required to register and adjust a view varies but typically takes minutes and increases as the number of views increases.
{The refinement with standard 3DGS takes $\sim 1$ hour.} During inference, we enjoy the same speed of \cite{3DGS} since we still leverage 3D Gaussians as our representation.} Please find more details in the supplementary.

\section{Results}
We provide evaluation details below, and conduct comparisons to other methods (Sec.~\ref{sec:experiment-comparison}), and an ablation study of the components of our algorithm (Sec.~\ref{sec:experiment-ablation-study}). We also encourage readers to look at the supplementary and accompanying video for more results.

\begin{table*}[t!]
    \caption{Quantitative evaluation on the Tanks\&Temples dataset. The best score is highlighted in bold throughout the paper. { $\dagger$: FSGS requires multi-view stereo estimation from COLMAP, which fails on 50$\%$ of total scenes. We therefore report the averaged metrics of the remaining scenes.}}
    \begin{tabular}{l|ccc|ccc|ccc}
    \hline
    \multirow{2}{*}{\textbf{Methods}} & \multicolumn{3}{c|}{\textbf{3 Views}}              & \multicolumn{3}{c|}{\textbf{6 Views}}              & \multicolumn{3}{c}{\textbf{12 Views}}             \\ \cline{2-10} 
    & \textbf{PSNR $\uparrow$}        & \textbf{SSIM $\uparrow$}        & \textbf{LPIPS $\downarrow$}        & \textbf{PSNR $\uparrow$}        & \textbf{SSIM $\uparrow$}        & \textbf{LPIPS $\downarrow$}        & \textbf{PSNR $\uparrow$}        & \textbf{SSIM $\uparrow$}        & \textbf{LPIPS $\downarrow$}       \\ \hline
    Instant-NGP~\faCamera      & 15.31 & 0.42 & 0.56 & 17.52 & 0.56 & 0.47 & 20.21 & 0.69 & 0.35 \\
    3DGS~\faCamera             & 15.21 & 0.46 & 0.43 & 20.17 & 0.71 & 0.24 & 23.60 & 0.81 & 0.17 \\
    FSGS$\dagger$~\faCamera & 19.23 & 0.58 & 0.37 & 23.55 & 0.74 & 0.28 & 26.81 & 0.83 & 0.22 \\
    GNT~\faCamera  & 17.80 & 0.57 & 0.29 & 22.52 & 0.77 & 0.18 & 24.56 & 0.82 & 0.14 \\
    LocalRF      & 16.06 & 0.49 & 0.70 & 16.31 & 0.50 & 0.67 & 18.68 & 0.54 & 0.61  \\
    NoPe-NeRF        & 12.05 & 0.35& 0.76 & 15.64 & 0.45& 0.65 & 18.12 & 0.49& 0.60\\
    CF 3DGS          & 14.91 & 0.43& 0.43 & 16.71 & 0.50& 0.41 & 18.62 & 0.59& 0.36\\ \hline
    Ours             & \textbf{20.37} & \textbf{0.66} & \textbf{0.26} & \textbf{25.18} & \textbf{0.81} & \textbf{0.16} & \textbf{28.65} & \textbf{0.88} & \textbf{0.10}\\ \hline
    \end{tabular}
    \label{table:tanks-temple}
\end{table*}

\begin{table*}[t!]
\caption{%
Quantitative evaluation on the Static Hikes dataset.}
\begin{tabular}{l|ccc|ccc|ccc}
\hline
\multirow{2}{*}{\textbf{Methods}} & \multicolumn{3}{c|}{\textbf{3 Views}}              & \multicolumn{3}{c|}{\textbf{6 Views}}              & \multicolumn{3}{c}{\textbf{12 Views}}             \\ \cline{2-10} 
& \textbf{PSNR $\uparrow$}        & \textbf{SSIM $\uparrow$}        & \textbf{LPIPS $\downarrow$}        & \textbf{PSNR $\uparrow$}        & \textbf{SSIM $\uparrow$}        & \textbf{LPIPS $\downarrow$}        & \textbf{PSNR $\uparrow$}        & \textbf{SSIM $\uparrow$}        & \textbf{LPIPS $\downarrow$}       \\ \hline
LocalRF        & 15.97 & 0.33 & 0.47 & 18.32 & 0.47 & 0.43 & \textbf{20.13} & \textbf{0.54} & 0.41 \\
NoPe-NeRF        & 14.85 & 0.25 & 0.67 & 18.59 & 0.34 & 0.57 & 18.19 & 0.34 & 0.59 \\
CF 3DGS          & 15.45 & 0.28 & 0.60 & 17.02 & 0.35 & 0.52 & 17.65 & 0.39 & 0.46 \\ \hline
Ours             & \textbf{16.35} & \textbf{0.38} & \textbf{0.37} & \textbf{18.96} & \textbf{0.50} & \textbf{0.31} & 19.70 & 0.53& \textbf{0.29} \\ \hline
\end{tabular}
\label{table:Hiking}
\end{table*}

\begin{table}[t]
\caption{PSNR Score $\uparrow$ on testing views for investigating the effects of number of training views (first row). %
}
\centering
\begin{tabular}{l|cccccc}
\hline
{\textbf{Methods}} & 3 \faCamera & 4 \faCamera & 6 \faCamera & 12 \faCamera & 24 \faCamera & 60 \faCamera \\ \hline
Instant-NGP~\faCamera                      & 16.46          & 16.94          & 17.19          & 17.86          & 18.57          & 21.10          \\
3DGS~\faCamera                             & 14.99          & 14.99          & 17.76          & 17.12          & 25.35          & 26.95          \\ \hline
Ours                             & \textbf{21.54} & \textbf{25.36} & \textbf{29.00} & \textbf{32.09} & \textbf{33.73} & \textbf{35.93} \\ \hline
\end{tabular}
\label{table:number_of_views}
\end{table}

\paragraph{Evaluation Details.}
{
To compare with methods which require initialized camera poses, we use SfM \cite{sfm1, sfm2} for registration. For fairness, SfM only sees training views for reconstruction, and registers testing views after reconstruction, in which bundle adjustment is not included. 
}

\paragraph{Datasets.} 
{We evaluate on datasets which contain non-trivial camera movements but ensure that there is overlap between consecutive frames, as has been done by other pose-free methods \cite{nopenerf, colmap-free, localrf} for evaluation.}
Following Bian et al.~\shortcite{nopenerf}, we use $8$ scenes of the \textbf{Tanks\&Temples} dataset \cite{tanks_and_temples}.
Following Meuleman et al.~\shortcite{localrf}, we also use $5$ scenes of the \textbf{Static Hikes} dataset \cite{localrf}. 
We estimate the monocular depth per frame with \cite{MariGold}, and we also show the sparse view synthesis results using another monocular depth estimator \cite{midas2} in the supplementary.
{For training, we select $n$ evenly distributed frames and use the others for testing. 
For example, when $n=3$, the first, middle and last frames are used for training. %
}

\subsection{Comparison}
\label{sec:experiment-comparison}
We compare with pose-free methods: COLMAP-Free 3DGS (CF 3DGS) \cite{colmap-free}, NoPe-NeRF \cite{nopenerf}, LocalRF \cite{localrf}; pose-required reconstruction methods: Instant-NGP \cite{instant-ngp}, 3DGS \cite{3DGS}; a pose-required sparse-view synthesis method: FSGS \cite{fsgs}; and a pose-required generalizable method: GNT \cite{GNT}.

\paragraph{Quantitative Evaluation. }
{We report the {averaged} evaluation results of testing views on {all 8 scenes of the} Tanks\&Temples dataset in Table~\ref{table:tanks-temple}.
We evaluate on the case $n=3, 6, 12$ and measure the difference between synthesized results and ground-truth images.
For the Static Hikes dataset, since SfM fails in many cases, we compare with pose-free methods only in Table~\ref{table:Hiking} {and report the average results of testing views on all 5 scenes}. Our method achieves the best metrics compared to other pose-free methods in most cases, and outperforms pose-required methods{, which is also shown in Fig.~\ref{fig:teaser}}.}{ Notice that the Static Hikes dataset comes with {stronger ambiguities and non-smooth camera trajectories}, resulting in relatively lower metrics of ours compared to those on the Tanks\&Temples dataset, which is analyzed in the supplementary. Even though LocalRF achieves higher metrics of PSNR and SSIM in terms of 12 views in the Static Hikes dataset, our method has much lower LPIPS metrics and performs better than it on the Tanks\&Temples dataset.}

{We also report the effect of the number of training views on the PSNR metric of testing views for the ``Horse'' scene ($120$ frames in total) in Table~\ref{table:number_of_views}. We compare with standard view synthesis methods: Instant-NGP and 3DGS. Our method achieves a high metric with sparse views and outperforms alternatives in all cases.} 

We also evaluate the performance of our method with respect to the ordering of training images. We experiment on the $3$ training images case of Tanks \& Temples dataset, and randomly shuffle the training images before feeding them into our sparse view synthesis pipeline. 
Since our pipeline relies on overlapping between two consecutive training images, random shuffling reduces the overlapping compared to the original order and therefore increases the difficulty of registration. In terms of testing views, the metrics of PSNR, SSIM, LPIPS change from $20.37$, $0.66$, and $0.26$ when using the original order into $19.06$, $0.59$, and $0.30$ when using randomly shuffled order. 

\begin{table}[t]
    \caption{PSNR Score $\uparrow$ on testing views for ablation models.}
    \centering
    \begin{tabular}{l|ccccc}
    \hline
    \textbf{Methods} & Config-A & Config-B & Config-C & Config-D & Ours                            \\ \hline
    \textbf{PSNR$\uparrow$} & 16.29    & 17.09    & 19.64    & 17.50    & \textbf{23.33} \\ \hline
    \end{tabular}
    \label{table:quantitative_ablation}
\end{table}

\begin{figure*}[t!]
  \includegraphics[width=0.9\linewidth]{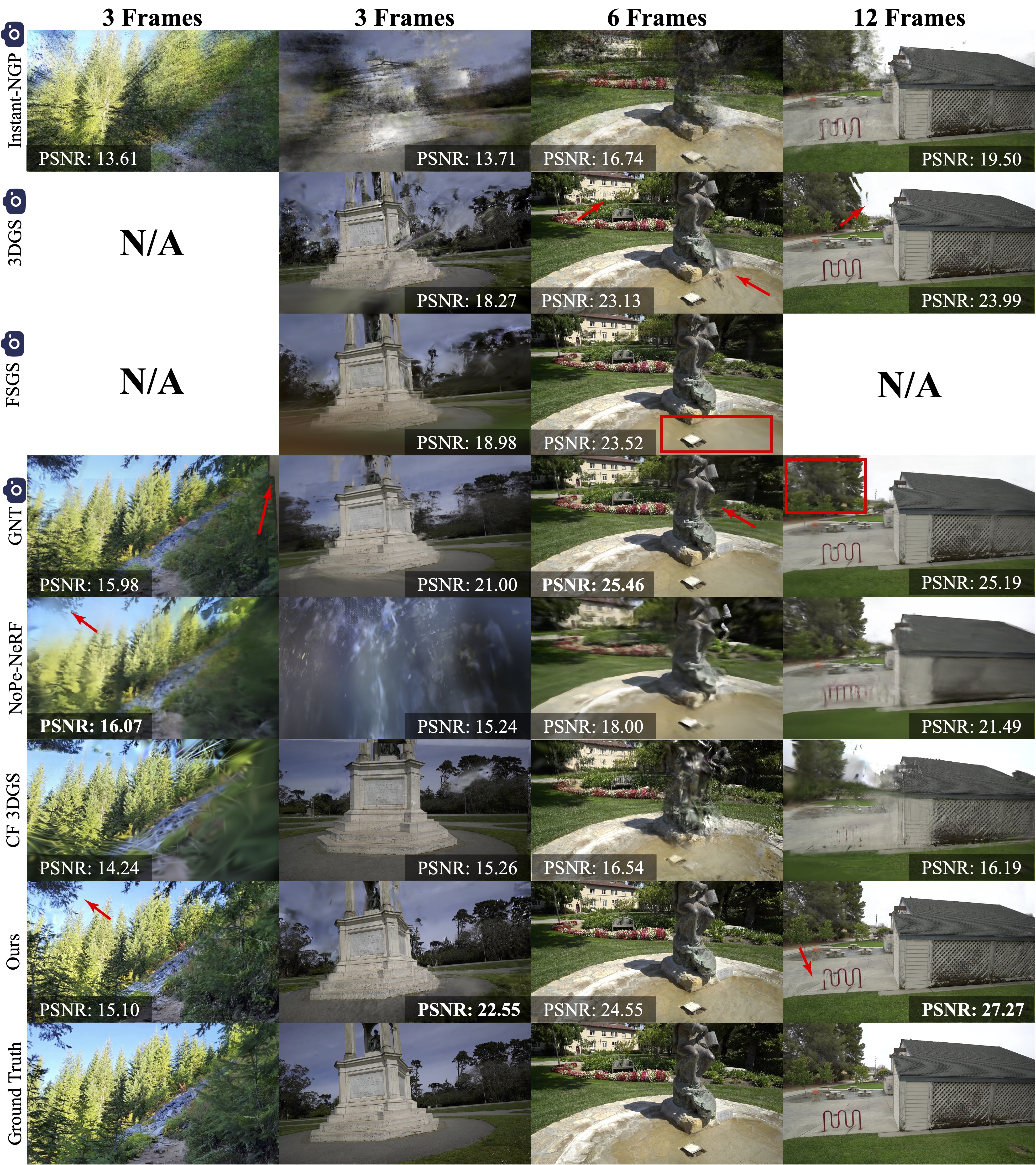}
  
  \caption{
  Qualitative comparison of different methods for sparse view synthesis. %
  From left to right, we use $3, 3, 6$ and $12$ frames as training views and others for testing. The scenes are, from the left to right: Forest from Static Hikes; Francis, Ignatius, Barn from Tanks\&Temples.
  Some subtle differences in quality are highlighted by arrows/rectangles. 
  {Since COLMAP fails for ``Forest'' (3 views) and multi-view stereo estimation fails for ``Barn'' (12 views), methods cannot handle these cases are denoted by ``N/A''. For Instant-NGP and GNT, we show their results on ``Forest'' (3 views) by giving ground-truth poses which are estimated given both training and testing views.
  }
  {In ``Forest'', due to its complexity (analyzed in the supplementary), our method cannot achieve high PSNR score despite much more visually pleasing and sharp results.}
  {In ``Francis'', Instant-NGP and 3DGS contain visible artifacts, while FSGS is too smooth on the grass. NoPe-NeRF and CF 3DGS cannot process sparse views well, leading to complete failure or misalignment.
  In ``Ignatius'', Instant-NGP, 3DGS and GNT contain blurry artifacts pointed out by arrows, despite slightly higher metric of GNT than ours. FSGS and NoPe-NeRF are overly smooth, while CF 3DGS cannot align the camera pose. 
  In ``Barn'', Instant-NGP, NoPe-NeRF and CF 3DGS cannot produce sharp rendered results, with blurs around red pillars. 3DGS cannot synthesize faithful background with missing trees and the telegraph pole. GNT synthesizes blurry trees. Our synthesized result also has certain artifacts around red pillars, but enjoys the best PSNR score.}
  Images credit by \citet{localrf} and \citet{tanks_and_temples}.
  }
  \label{fig:qualitative}
\end{figure*}

\paragraph{Qualitative Evaluation. }
{We show novel views synthesized from sparse inputs on{ one scene in Fig.~\ref{fig:teaser} and} four scenes in Fig.~\ref{fig:qualitative}.
Fitting a solution from scratch or sparse initialization is ambiguous, resulting in noisy and blurry {backgrounds in ``Forest'', ``Family'', ``Ignatius'', ``Francis'' and ``Barn''}, which also lose high-frequency details {such as words under the statue in ``Family''} in Instant-NGP, NoPe-NeRF, 3DGS, and FSGS. Besides, NoPe-NeRF, GNT, and CF 3DGS fail to handle sparse training views{, resulting in the image mismatch in ``Francis''}. In comparison, our method achieves better results in terms of both synthesis quality and pose alignment.}

\begin{figure*}[t!]
  \centering
  \includegraphics[width=0.925\linewidth]{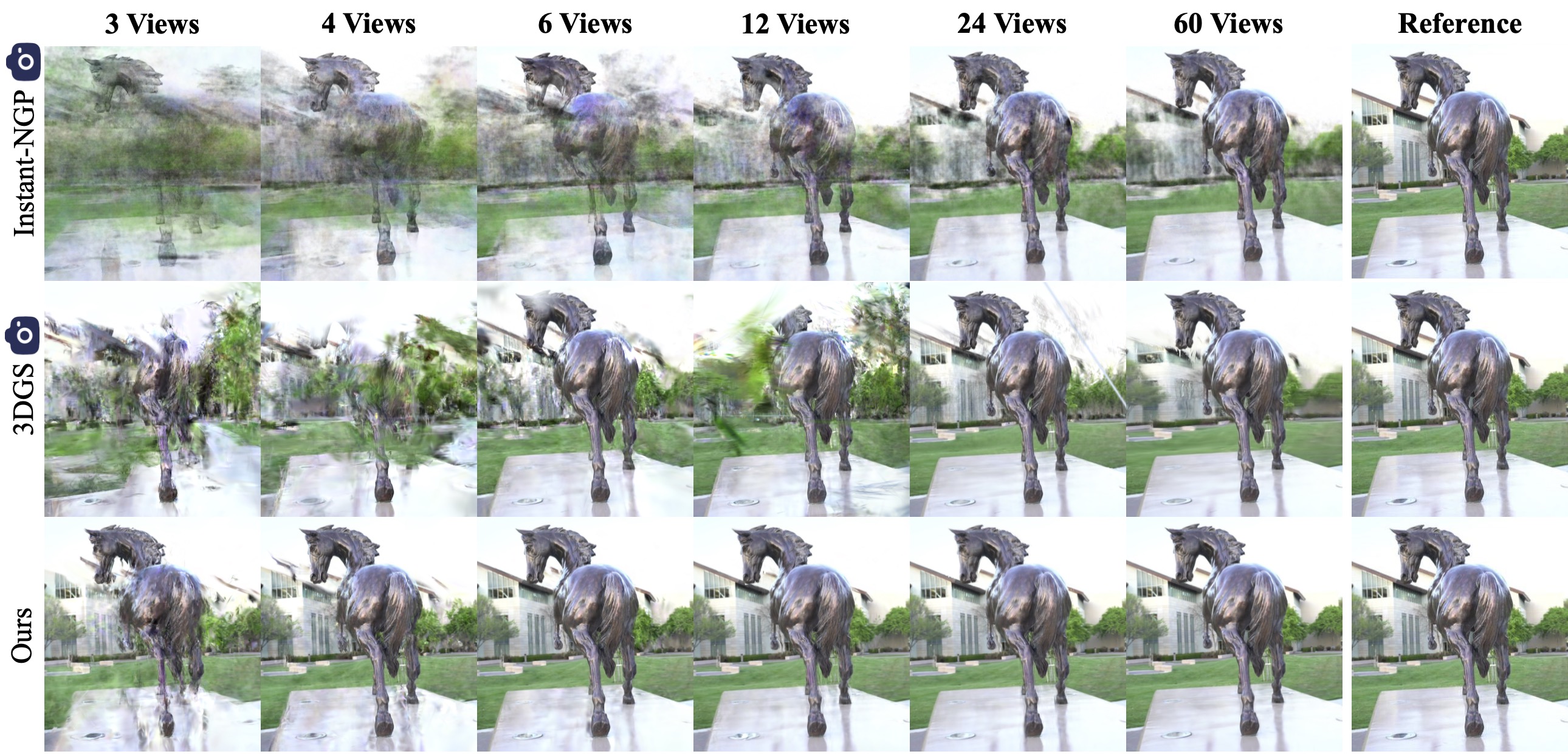}
  
  \caption{
  Qualitative comparison of the effects of the number of training views. For the ``Horse'' scene in Tanks\&Temples ($120$ frames in total), from left to right, we use $3, 4, 6, 12, 24$ and $60$ frames as training views and compare the results on the same testing view. {Using only $3$ views, our results are still somewhat ambiguous, but} our method can synthesize faithful results with 4 views, and they improve in accuracy with more views. In comparison, other methods struggle to synthesize faithful results until $60$ views, where the background is still not accurate. {Instant-NGP features blurry artifacts, while 3DGS features high-frequency artifacts, resulting in occasional failure as in the $12$ views case and lower PSNR at very sparse views in Table~\ref{table:number_of_views}.}
  Image credits by \citet{tanks_and_temples}.
  }
  \label{fig:number-of-views}
\end{figure*}

{We investigate the effects of the number of training views in Fig.~\ref{fig:number-of-views}. We compare with standard pose-required view synthesis methods: Instant-NGP and 3DGS. Other methods struggle to synthesize high-quality novel view results, with blurry or high-frequency artifacts. Our results are still ambiguous with only 3 views, but are greatly improved with 4 and more views.}

\subsection{Ablation Study}
\label{sec:experiment-ablation-study}
We compare with various baselines to validate our proposed algorithm {for achieving a coarse solution. They are all passed through a low-pass filter and refined using the standard method \cite{3DGS}.}
To validate the correspondence-based optimization, we set $\lambda_1=0$ in Eqn.~\ref{eqn:optimization-objective}, denoted as ``Config-A''.
To validate the proposed differentiable approximate surface rendering, we replace the proposed rendering with the one in \cite{depth-gs} and its extension to surface points as ``Config-B''. 
To validate the adjustment which aligns the monocular depth, we {skip the adjustment step in the Fig.~\ref{fig:pipeline} (c) as ``Config-C''.}
{Besides, we }directly back-project pixels into the scene with the original monocular depth and camera poses estimated by SfM, and denote this as ``Config-D''. 

\begin{figure*}[t!]
  \centering
  \includegraphics[width=0.925\linewidth]{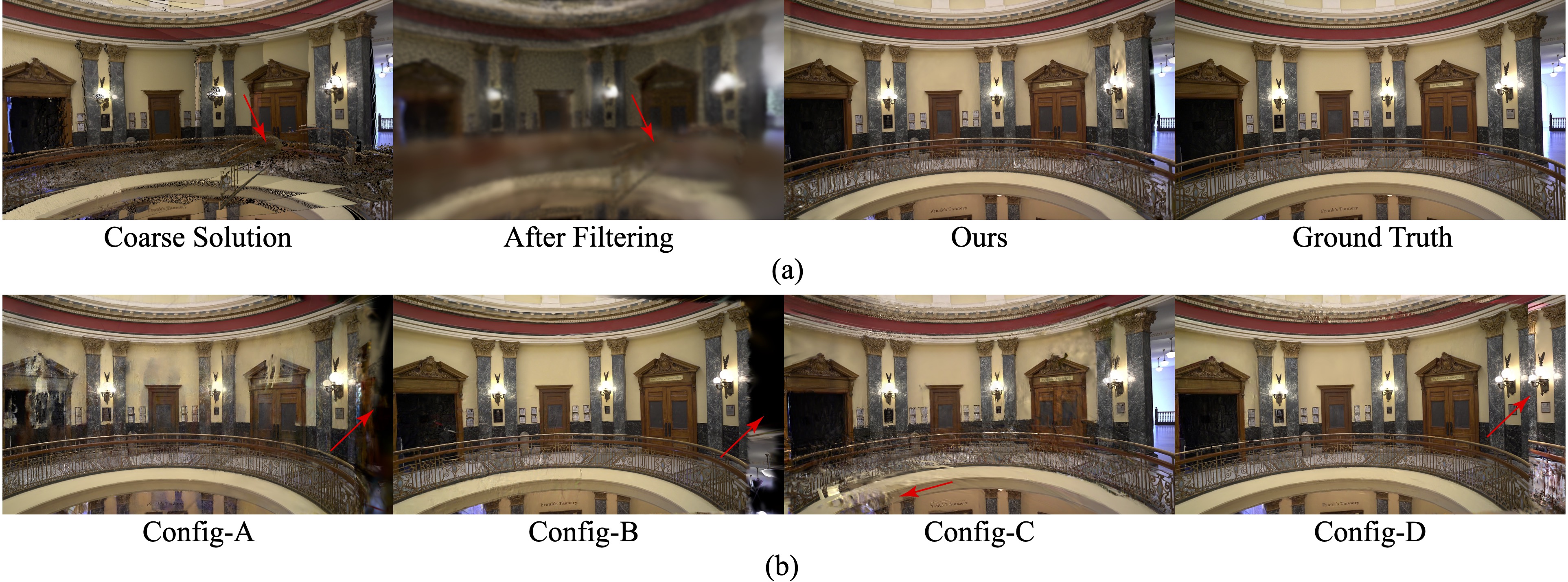}
  
  \caption{
  {Given the ``Museum'' scene in Tanks\&Temples ($100$ frames in total), we use $6$ frames as training views and others for testing. We show the synthesized results on the same testing view for different configurations. Regions of interest are emphasized by arrows.
  (a) We show the effects of applying a low-pass filter and refinement. As pointed by the arrow, even though we manage to align the monocular depths, coarse solution still has high-frequency errors due to inaccuracy of monocular depths. After applying a low-pass filter, error-prune high-frequency information is removed and then faithfully reproduced by refinement. (b) We show the results of different ablation models. The ground-truth is the same with that in (a). ``Config-A'' and ``Config-B'' cannot register the camera poses successfully. Therefore, when the testing pose deviates from training poses, there are missing regions, pointed out by arrows. ``Config-C'' and ``Config-D'' cannot ensure the alignment between camera poses and monocular depths, leading to artifacts in ``Config-C'' and wrong synthesized results in ``Config-D'', pointed out by arrows.
  }
  Image credits by \citet{tanks_and_temples}.
  }
  \label{fig:ablation-study}
\end{figure*}

{We use the ``Museum'' scene in Tanks\&Temples and evenly select $6$ frames as the training views, with the other $94$ frames as testing views for evaluation. We report metrics in Table~\ref{table:quantitative_ablation}. Our full model achieves the best metrics. 
{We show the effects of applying low-pass filtering and refinement in Fig.~\ref{fig:ablation-study} (a) and }
qualitative comparison of the different configurations in Fig.~\ref{fig:ablation-study} (b). }{The synthesis quality benefits from refinement. Moreover, }in comparison, ``Config-A'', which does not leverage correspondences, and ``Config-B'' cannot register camera poses successfully, resulting in missing regions. Note that correspondence detection does not fail, but the camera optimization fails in ``Config-B''. ``Config-C'' and ``Config-D'' cannot ensure the alignment between camera poses and monocular depths, resulting in sub-optimal synthesized results.
\section{Conclusions, and Future Work}
Sparse {view synthesis} is desirable but challenging due to insufficient camera coverage. 
From the perspective of fitting the signal, the problem is still very ambiguous for {under-sampled views} despite introducing certain priors, such as depth.
{Thanks to the explicitness of Gaussian splatting, }we propose {to construct a coarse solution, where optimization is still involved, for sparse view synthesis.} 
{It is then refined for faithful high-frequency details.}
To effectively reach a coarse solution, we propose to unify registration and adjustment in a fully differentiable pipeline, which leverages long-range information to address the limitation of pixel-wise supervision. A differentiable approximation of {the expected} surface in Gaussian splatting is also proposed for optimization.

Our method is not without limitations. {We can achieve reasonable but not perfect sparse view synthesis for too few training views.} The construction of the coarse solution depends on the scale-consistent assumption of estimated monocular depth, which does not hold for complex scenes, such as $360^{\circ}$ scenes. By assuming overlapping between consecutive frames, our method also cannot handle unordered collection of images well. Besides, since a Gaussian kernel does not necessarily correspond to a valid surface{, a more accurate definition of the surface is welcome. }
In the future, it would be interesting to explore how to {adjust monocular depths more accurately and incorporate novel view {constraints} to enhance view synthesis quality, and extend our method to unordered collections of images. }
{In conclusion, our work {proposes to construct a solution with correspondence-based optimization instead of fitting one from scratch} to solve sparse view synthesis {without camera poses}. {We} achieve significantly better results {than other pose-free methods and even outperform methods which rely on off-the-shelf estimated camera poses}. This {framework} paves the way for future study on sparse view synthesis, few-shot reconstruction, and {reconstruction without camera poses}.}

\begin{acks}
This work was supported in part by the Intelligence Advanced Research Projects Activity (IARPA) via Department of Interior/ Interior Business Center (DOI/IBC) contract number 140D0423C0076. The U.S. Government is authorized to reproduce and distribute reprints for Governmental purposes notwithstanding any copyright annotation thereon. Disclaimer: The views and conclusions contained herein are those of the authors and should not be interpreted as necessarily representing the official policies or endorsements, either expressed or implied, of IARPA, DOI/IBC, or the U.S. Government. We also acknowledge support from NSF grants 2100237 and 2120019 for the Nautilus cluster, gifts from Adobe, Google, Qualcomm and Rembrand, the Ronald L. Graham Chair and the UC San Diego Center for Visual Computing.
 
\end{acks}

\bibliographystyle{ACM-Reference-Format}
\bibliography{bibliography}


\begin{thebibliography}{53}


\ifx \showCODEN    \undefined \def \showCODEN     #1{\unskip}     \fi
\ifx \showDOI      \undefined \def \showDOI       #1{#1}\fi
\ifx \showISBNx    \undefined \def \showISBNx     #1{\unskip}     \fi
\ifx \showISBNxiii \undefined \def \showISBNxiii  #1{\unskip}     \fi
\ifx \showISSN     \undefined \def \showISSN      #1{\unskip}     \fi
\ifx \showLCCN     \undefined \def \showLCCN      #1{\unskip}     \fi
\ifx \shownote     \undefined \def \shownote      #1{#1}          \fi
\ifx \showarticletitle \undefined \def \showarticletitle #1{#1}   \fi
\ifx \showURL      \undefined \def \showURL       {\relax}        \fi
\providecommand\bibfield[2]{#2}
\providecommand\bibinfo[2]{#2}
\providecommand\natexlab[1]{#1}
\providecommand\showeprint[2][]{arXiv:#2}

\bibitem[Barron et~al\mbox{.}(2023)]%
        {zip-nerf}
\bibfield{author}{\bibinfo{person}{Jonathan~T. Barron}, \bibinfo{person}{Ben
  Mildenhall}, \bibinfo{person}{Dor Verbin}, \bibinfo{person}{Pratul~P.
  Srinivasan}, {and} \bibinfo{person}{Peter Hedman}.}
  \bibinfo{year}{2023}\natexlab{}.
\newblock \showarticletitle{Zip-NeRF: Anti-Aliased Grid-Based Neural Radiance
  Fields}.
\newblock \bibinfo{journal}{\emph{ICCV}} (\bibinfo{year}{2023}).
\newblock


\bibitem[Bian et~al\mbox{.}(2023)]%
        {nopenerf}
\bibfield{author}{\bibinfo{person}{Wenjing Bian}, \bibinfo{person}{Zirui Wang},
  \bibinfo{person}{Kejie Li}, \bibinfo{person}{Jiawang Bian}, {and}
  \bibinfo{person}{Victor~Adrian Prisacariu}.} \bibinfo{year}{2023}\natexlab{}.
\newblock \showarticletitle{NoPe-NeRF: Optimising Neural Radiance Field with No
  Pose Prior}.
\newblock \bibinfo{journal}{\emph{CVPR}}.
\newblock


\bibitem[Birkl et~al\mbox{.}(2023)]%
        {midas2}
\bibfield{author}{\bibinfo{person}{Reiner Birkl}, \bibinfo{person}{Diana Wofk},
  {and} \bibinfo{person}{Matthias M{\"u}ller}.}
  \bibinfo{year}{2023}\natexlab{}.
\newblock \showarticletitle{MiDaS v3.1 -- A Model Zoo for Robust Monocular
  Relative Depth Estimation}.
\newblock \bibinfo{journal}{\emph{arXiv preprint arXiv:2307.14460}}
  (\bibinfo{year}{2023}).
\newblock


\bibitem[Chai et~al\mbox{.}(2000)]%
        {related_classic_work_3}
\bibfield{author}{\bibinfo{person}{Jin-Xiang Chai}, \bibinfo{person}{Xin Tong},
  \bibinfo{person}{Shing-Chow Chan}, {and} \bibinfo{person}{Heung-Yeung Shum}.}
  \bibinfo{year}{2000}\natexlab{}.
\newblock \showarticletitle{Plenoptic Sampling}. In
  \bibinfo{booktitle}{\emph{Proceedings of the 27th Annual Conference on
  Computer Graphics and Interactive Techniques}}
  \emph{(\bibinfo{series}{SIGGRAPH '00})}. \bibinfo{publisher}{ACM
  Press/Addison-Wesley Publishing Co.}, \bibinfo{address}{USA},
  \bibinfo{pages}{307–318}.
\newblock


\bibitem[Chen and Williams(1993)]%
        {related_classic_work_1}
\bibfield{author}{\bibinfo{person}{Shenchang~Eric Chen} {and}
  \bibinfo{person}{Lance Williams}.} \bibinfo{year}{1993}\natexlab{}.
\newblock \showarticletitle{View interpolation for image synthesis}. In
  \bibinfo{booktitle}{\emph{Proceedings of the 20th Annual Conference on
  Computer Graphics and Interactive Techniques}} (Anaheim, CA)
  \emph{(\bibinfo{series}{SIGGRAPH '93})}. \bibinfo{publisher}{Association for
  Computing Machinery}, \bibinfo{address}{New York, NY, USA},
  \bibinfo{pages}{279–288}.
\newblock
\showISBNx{0897916018}


\bibitem[Chung et~al\mbox{.}(2023)]%
        {depth-gs}
\bibfield{author}{\bibinfo{person}{Jaeyoung Chung}, \bibinfo{person}{Jeongtaek
  Oh}, {and} \bibinfo{person}{Kyoung~Mu Lee}.} \bibinfo{year}{2023}\natexlab{}.
\newblock \showarticletitle{Depth-Regularized Optimization for 3D Gaussian
  Splatting in Few-Shot Images}.
\newblock \bibinfo{journal}{\emph{arXiv preprint arXiv:2311.13398}}
  (\bibinfo{year}{2023}).
\newblock


\bibitem[Deng et~al\mbox{.}(2022)]%
        {DSNeRF}
\bibfield{author}{\bibinfo{person}{Kangle Deng}, \bibinfo{person}{Andrew Liu},
  \bibinfo{person}{Jun{-}Yan Zhu}, {and} \bibinfo{person}{Deva Ramanan}.}
  \bibinfo{year}{2022}\natexlab{}.
\newblock \showarticletitle{Depth-supervised NeRF: Fewer Views and Faster
  Training for Free}. In \bibinfo{booktitle}{\emph{{IEEE/CVF} Conference on
  Computer Vision and Pattern Recognition, {CVPR} 2022, New Orleans, LA, USA,
  June 18-24, 2022}}. \bibinfo{publisher}{{IEEE}},
  \bibinfo{pages}{12872--12881}.
\newblock


\bibitem[Fu et~al\mbox{.}(2023)]%
        {colmap-free}
\bibfield{author}{\bibinfo{person}{Yang Fu}, \bibinfo{person}{Sifei Liu},
  \bibinfo{person}{Amey Kulkarni}, \bibinfo{person}{Jan Kautz},
  \bibinfo{person}{Alexei~A Efros}, {and} \bibinfo{person}{Xiaolong Wang}.}
  \bibinfo{year}{2023}\natexlab{}.
\newblock \showarticletitle{COLMAP-Free 3D Gaussian Splatting}.
\newblock \bibinfo{journal}{\emph{arXiv preprint arXiv:2312.07504}}
  (\bibinfo{year}{2023}).
\newblock


\bibitem[Gortler et~al\mbox{.}(1996)]%
        {related_classic_work_7}
\bibfield{author}{\bibinfo{person}{Steven~J. Gortler}, \bibinfo{person}{Radek
  Grzeszczuk}, \bibinfo{person}{Richard Szeliski}, {and}
  \bibinfo{person}{Michael~F. Cohen}.} \bibinfo{year}{1996}\natexlab{}.
\newblock \showarticletitle{The lumigraph}. In
  \bibinfo{booktitle}{\emph{Proceedings of the 23rd Annual Conference on
  Computer Graphics and Interactive Techniques}}
  \emph{(\bibinfo{series}{SIGGRAPH '96})}. \bibinfo{publisher}{Association for
  Computing Machinery}, \bibinfo{address}{New York, NY, USA},
  \bibinfo{pages}{43–54}.
\newblock
\showISBNx{0897917464}


\bibitem[Gu{\'e}don and Lepetit(2023)]%
        {sugar}
\bibfield{author}{\bibinfo{person}{Antoine Gu{\'e}don} {and}
  \bibinfo{person}{Vincent Lepetit}.} \bibinfo{year}{2023}\natexlab{}.
\newblock \showarticletitle{SuGaR: Surface-Aligned Gaussian Splatting for
  Efficient 3D Mesh Reconstruction and High-Quality Mesh Rendering}.
\newblock \bibinfo{journal}{\emph{arXiv preprint arXiv:2311.12775}}
  (\bibinfo{year}{2023}).
\newblock


\bibitem[Hedman and Kopf(2018)]%
        {instant-3d-photography}
\bibfield{author}{\bibinfo{person}{Peter Hedman} {and}
  \bibinfo{person}{Johannes Kopf}.} \bibinfo{year}{2018}\natexlab{}.
\newblock \showarticletitle{Instant 3D photography}.
\newblock \bibinfo{journal}{\emph{ACM Trans. Graph.}} \bibinfo{volume}{37},
  \bibinfo{number}{4}, Article \bibinfo{articleno}{101} (\bibinfo{date}{jul}
  \bibinfo{year}{2018}), \bibinfo{numpages}{12}~pages.
\newblock


\bibitem[Heo et~al\mbox{.}(2023)]%
        {robust_refinement}
\bibfield{author}{\bibinfo{person}{Hwan Heo}, \bibinfo{person}{Taekyung Kim},
  \bibinfo{person}{Jiyoung Lee}, \bibinfo{person}{Jaewon Lee},
  \bibinfo{person}{Soohyun Kim}, \bibinfo{person}{Hyunwoo~J Kim}, {and}
  \bibinfo{person}{Jin-Hwa Kim}.} \bibinfo{year}{2023}\natexlab{}.
\newblock \showarticletitle{Robust camera pose refinement for multi-resolution
  hash encoding}. In \bibinfo{booktitle}{\emph{International Conference on
  Machine Learning}}. PMLR, \bibinfo{pages}{13000--13016}.
\newblock


\bibitem[Kalantari et~al\mbox{.}(2016)]%
        {related_classic_work_5}
\bibfield{author}{\bibinfo{person}{Nima~Khademi Kalantari},
  \bibinfo{person}{Ting-Chun Wang}, {and} \bibinfo{person}{Ravi Ramamoorthi}.}
  \bibinfo{year}{2016}\natexlab{}.
\newblock \showarticletitle{Learning-Based View Synthesis for Light Field
  Cameras}.
\newblock \bibinfo{journal}{\emph{ACM Trans. Graph.}} \bibinfo{volume}{35},
  \bibinfo{number}{6}, Article \bibinfo{articleno}{193} (\bibinfo{year}{2016}),
  \bibinfo{numpages}{10}~pages.
\newblock


\bibitem[Ke et~al\mbox{.}(2023)]%
        {MariGold}
\bibfield{author}{\bibinfo{person}{Bingxin Ke}, \bibinfo{person}{Anton
  Obukhov}, \bibinfo{person}{Shengyu Huang}, \bibinfo{person}{Nando Metzger},
  \bibinfo{person}{Rodrigo~Caye Daudt}, {and} \bibinfo{person}{Konrad
  Schindler}.} \bibinfo{year}{2023}\natexlab{}.
\newblock \bibinfo{title}{Repurposing Diffusion-Based Image Generators for
  Monocular Depth Estimation}.
\newblock
\newblock
\showeprint[arxiv]{2312.02145}~[cs.CV]


\bibitem[Keetha et~al\mbox{.}(2023)]%
        {splatam}
\bibfield{author}{\bibinfo{person}{Nikhil Keetha}, \bibinfo{person}{Jay
  Karhade}, \bibinfo{person}{Krishna~Murthy Jatavallabhula},
  \bibinfo{person}{Gengshan Yang}, \bibinfo{person}{Sebastian Scherer},
  \bibinfo{person}{Deva Ramanan}, {and} \bibinfo{person}{Jonathon Luiten}.}
  \bibinfo{year}{2023}\natexlab{}.
\newblock \showarticletitle{SplaTAM: Splat, Track \& Map 3D Gaussians for Dense
  RGB-D SLAM}.
\newblock \bibinfo{journal}{\emph{arXiv preprint}} (\bibinfo{year}{2023}).
\newblock


\bibitem[Kerbl et~al\mbox{.}(2023)]%
        {3DGS}
\bibfield{author}{\bibinfo{person}{Bernhard Kerbl}, \bibinfo{person}{Georgios
  Kopanas}, \bibinfo{person}{Thomas Leimk{\"u}hler}, {and}
  \bibinfo{person}{George Drettakis}.} \bibinfo{year}{2023}\natexlab{}.
\newblock \showarticletitle{3D Gaussian Splatting for Real-Time Radiance Field
  Rendering}.
\newblock \bibinfo{journal}{\emph{ACM Transactions on Graphics}}
  \bibinfo{volume}{42}, \bibinfo{number}{4} (\bibinfo{date}{July}
  \bibinfo{year}{2023}).
\newblock


\bibitem[Kim et~al\mbox{.}(2022)]%
        {info-nerf}
\bibfield{author}{\bibinfo{person}{Mijeong Kim}, \bibinfo{person}{Seonguk Seo},
  {and} \bibinfo{person}{Bohyung Han}.} \bibinfo{year}{2022}\natexlab{}.
\newblock \showarticletitle{InfoNeRF: Ray Entropy Minimization for Few-Shot
  Neural Volume Rendering}. In \bibinfo{booktitle}{\emph{{IEEE/CVF} Conference
  on Computer Vision and Pattern Recognition, {CVPR} 2022, New Orleans, LA,
  USA, June 18-24, 2022}}. \bibinfo{publisher}{{IEEE}},
  \bibinfo{pages}{12902--12911}.
\newblock


\bibitem[Knapitsch et~al\mbox{.}(2017)]%
        {tanks_and_temples}
\bibfield{author}{\bibinfo{person}{Arno Knapitsch}, \bibinfo{person}{Jaesik
  Park}, \bibinfo{person}{Qian-Yi Zhou}, {and} \bibinfo{person}{Vladlen
  Koltun}.} \bibinfo{year}{2017}\natexlab{}.
\newblock \showarticletitle{Tanks and Temples: Benchmarking Large-Scale Scene
  Reconstruction}.
\newblock \bibinfo{journal}{\emph{ACM Transactions on Graphics}}
  \bibinfo{volume}{36}, \bibinfo{number}{4} (\bibinfo{year}{2017}).
\newblock


\bibitem[Kopf et~al\mbox{.}(2021)]%
        {consistent_video_depth}
\bibfield{author}{\bibinfo{person}{Johannes Kopf}, \bibinfo{person}{Xuejian
  Rong}, {and} \bibinfo{person}{Jia-Bin Huang}.}
  \bibinfo{year}{2021}\natexlab{}.
\newblock \showarticletitle{Robust Consistent Video Depth Estimation}. In
  \bibinfo{booktitle}{\emph{IEEE/CVF Conference on Computer Vision and Pattern
  Recognition}}.
\newblock


\bibitem[Lassner and Zollhöfer(2021)]%
        {pulsar}
\bibfield{author}{\bibinfo{person}{Christoph Lassner} {and}
  \bibinfo{person}{Michael Zollhöfer}.} \bibinfo{year}{2021}\natexlab{}.
\newblock \showarticletitle{Pulsar: Efficient sphere-based neural rendering}.
  In \bibinfo{booktitle}{\emph{Proceedings of the IEEE/CVF Conference on
  Computer Vision and Pattern Recognition}}. \bibinfo{pages}{1440--1449}.
\newblock


\bibitem[Lepetit et~al\mbox{.}(2009)]%
        {epnp}
\bibfield{author}{\bibinfo{person}{Vincent Lepetit}, \bibinfo{person}{Francesc
  Moreno-Noguer}, {and} \bibinfo{person}{Pascal Fua}.}
  \bibinfo{year}{2009}\natexlab{}.
\newblock \showarticletitle{EPnP: An Accurate O(n) Solution to the PnP
  Problem}.
\newblock \bibinfo{journal}{\emph{International Journal Of Computer Vision}}
  \bibinfo{volume}{81} (\bibinfo{year}{2009}), \bibinfo{pages}{155--166}.
\newblock


\bibitem[Levoy and Hanrahan(1996)]%
        {related_classic_work_2}
\bibfield{author}{\bibinfo{person}{Marc Levoy} {and} \bibinfo{person}{Pat
  Hanrahan}.} \bibinfo{year}{1996}\natexlab{}.
\newblock \showarticletitle{Light field rendering}. In
  \bibinfo{booktitle}{\emph{Proceedings of the 23rd Annual Conference on
  Computer Graphics and Interactive Techniques}}
  \emph{(\bibinfo{series}{SIGGRAPH '96})}. \bibinfo{publisher}{Association for
  Computing Machinery}, \bibinfo{address}{New York, NY, USA},
  \bibinfo{pages}{31–42}.
\newblock
\showISBNx{0897917464}


\bibitem[Lin et~al\mbox{.}(2021)]%
        {barf}
\bibfield{author}{\bibinfo{person}{Chen-Hsuan Lin}, \bibinfo{person}{Wei-Chiu
  Ma}, \bibinfo{person}{Antonio Torralba}, {and} \bibinfo{person}{Simon
  Lucey}.} \bibinfo{year}{2021}\natexlab{}.
\newblock \showarticletitle{Barf: Bundle-adjusting neural radiance fields}. In
  \bibinfo{booktitle}{\emph{Proceedings of the IEEE/CVF International
  Conference on Computer Vision}}. \bibinfo{pages}{5741--5751}.
\newblock


\bibitem[McMillan and Bishop(1995)]%
        {related_classic_work_6}
\bibfield{author}{\bibinfo{person}{Leonard McMillan} {and}
  \bibinfo{person}{Gary Bishop}.} \bibinfo{year}{1995}\natexlab{}.
\newblock \showarticletitle{Plenoptic modeling: an image-based rendering
  system}. In \bibinfo{booktitle}{\emph{Proceedings of the 22nd Annual
  Conference on Computer Graphics and Interactive Techniques}}
  \emph{(\bibinfo{series}{SIGGRAPH '95})}. \bibinfo{publisher}{Association for
  Computing Machinery}, \bibinfo{address}{New York, NY, USA},
  \bibinfo{pages}{39–46}.
\newblock


\bibitem[Meng et~al\mbox{.}(2021)]%
        {gnerf}
\bibfield{author}{\bibinfo{person}{Quan Meng}, \bibinfo{person}{Anpei Chen},
  \bibinfo{person}{Haimin Luo}, \bibinfo{person}{Minye Wu},
  \bibinfo{person}{Hao Su}, \bibinfo{person}{Lan Xu}, \bibinfo{person}{Xuming
  He}, {and} \bibinfo{person}{Jingyi Yu}.} \bibinfo{year}{2021}\natexlab{}.
\newblock \showarticletitle{Gnerf: Gan-based neural radiance field without
  posed camera}. In \bibinfo{booktitle}{\emph{Proceedings of the IEEE/CVF
  International Conference on Computer Vision}}. \bibinfo{pages}{6351--6361}.
\newblock


\bibitem[Meuleman et~al\mbox{.}(2023)]%
        {localrf}
\bibfield{author}{\bibinfo{person}{Andreas Meuleman}, \bibinfo{person}{Yu-Lun
  Liu}, \bibinfo{person}{Chen Gao}, \bibinfo{person}{Jia-Bin Huang},
  \bibinfo{person}{Changil Kim}, \bibinfo{person}{Min~H. Kim}, {and}
  \bibinfo{person}{Johannes Kopf}.} \bibinfo{year}{2023}\natexlab{}.
\newblock \showarticletitle{Progressively Optimized Local Radiance Fields for
  Robust View Synthesis}. In \bibinfo{booktitle}{\emph{CVPR}}.
\newblock


\bibitem[Mildenhall et~al\mbox{.}(2019)]%
        {llff}
\bibfield{author}{\bibinfo{person}{Ben Mildenhall}, \bibinfo{person}{Pratul~P.
  Srinivasan}, \bibinfo{person}{Rodrigo Ortiz-Cayon},
  \bibinfo{person}{Nima~Khademi Kalantari}, \bibinfo{person}{Ravi Ramamoorthi},
  \bibinfo{person}{Ren Ng}, {and} \bibinfo{person}{Abhishek Kar}.}
  \bibinfo{year}{2019}\natexlab{}.
\newblock \showarticletitle{Local Light Field Fusion: Practical View Synthesis
  with Prescriptive Sampling Guidelines}.
\newblock \bibinfo{journal}{\emph{ACM Transactions on Graphics (TOG)}}
  (\bibinfo{year}{2019}).
\newblock


\bibitem[Mildenhall et~al\mbox{.}(2020)]%
        {nerf2020}
\bibfield{author}{\bibinfo{person}{Ben Mildenhall}, \bibinfo{person}{Pratul~P.
  Srinivasan}, \bibinfo{person}{Matthew Tancik}, \bibinfo{person}{Jonathan~T.
  Barron}, \bibinfo{person}{Ravi Ramamoorthi}, {and} \bibinfo{person}{Ren Ng}.}
  \bibinfo{year}{2020}\natexlab{}.
\newblock \showarticletitle{NeRF: Representing Scenes as Neural Radiance Fields
  for View Synthesis}. In \bibinfo{booktitle}{\emph{ECCV}}.
\newblock


\bibitem[Mildenhall et~al\mbox{.}(2022)]%
        {nerf}
\bibfield{author}{\bibinfo{person}{Ben Mildenhall}, \bibinfo{person}{Pratul~P.
  Srinivasan}, \bibinfo{person}{Matthew Tancik}, \bibinfo{person}{Jonathan~T.
  Barron}, \bibinfo{person}{Ravi Ramamoorthi}, {and} \bibinfo{person}{Ren Ng}.}
  \bibinfo{year}{2022}\natexlab{}.
\newblock \showarticletitle{NeRF: representing scenes as neural radiance fields
  for view synthesis}.
\newblock \bibinfo{journal}{\emph{Commun. {ACM}}} \bibinfo{volume}{65},
  \bibinfo{number}{1} (\bibinfo{year}{2022}), \bibinfo{pages}{99--106}.
\newblock


\bibitem[M{\"{u}}ller et~al\mbox{.}(2022)]%
        {instant-ngp}
\bibfield{author}{\bibinfo{person}{Thomas M{\"{u}}ller}, \bibinfo{person}{Alex
  Evans}, \bibinfo{person}{Christoph Schied}, {and} \bibinfo{person}{Alexander
  Keller}.} \bibinfo{year}{2022}\natexlab{}.
\newblock \showarticletitle{Instant neural graphics primitives with a
  multiresolution hash encoding}.
\newblock \bibinfo{journal}{\emph{{ACM} Trans. Graph.}} \bibinfo{volume}{41},
  \bibinfo{number}{4} (\bibinfo{year}{2022}), \bibinfo{pages}{102:1--102:15}.
\newblock


\bibitem[Niemeyer et~al\mbox{.}(2022)]%
        {reg-nerf}
\bibfield{author}{\bibinfo{person}{Michael Niemeyer},
  \bibinfo{person}{Jonathan~T. Barron}, \bibinfo{person}{Ben Mildenhall},
  \bibinfo{person}{Mehdi S.~M. Sajjadi}, \bibinfo{person}{Andreas Geiger},
  {and} \bibinfo{person}{Noha Radwan}.} \bibinfo{year}{2022}\natexlab{}.
\newblock \showarticletitle{RegNeRF: Regularizing Neural Radiance Fields for
  View Synthesis from Sparse Inputs}. In \bibinfo{booktitle}{\emph{{IEEE/CVF}
  Conference on Computer Vision and Pattern Recognition, {CVPR} 2022, New
  Orleans, LA, USA, June 18-24, 2022}}. \bibinfo{publisher}{{IEEE}},
  \bibinfo{pages}{5470--5480}.
\newblock


\bibitem[Park et~al\mbox{.}(2023)]%
        {camp}
\bibfield{author}{\bibinfo{person}{Keunhong Park}, \bibinfo{person}{Philipp
  Henzler}, \bibinfo{person}{Ben Mildenhall}, \bibinfo{person}{Jonathan~T
  Barron}, {and} \bibinfo{person}{Ricardo Martin-Brualla}.}
  \bibinfo{year}{2023}\natexlab{}.
\newblock \showarticletitle{CamP: Camera preconditioning for neural radiance
  fields}.
\newblock \bibinfo{journal}{\emph{ACM Transactions on Graphics (TOG)}}
  \bibinfo{volume}{42}, \bibinfo{number}{6} (\bibinfo{year}{2023}),
  \bibinfo{pages}{1--11}.
\newblock


\bibitem[Ranftl et~al\mbox{.}(2022)]%
        {midas1}
\bibfield{author}{\bibinfo{person}{Ren\'{e} Ranftl}, \bibinfo{person}{Katrin
  Lasinger}, \bibinfo{person}{David Hafner}, \bibinfo{person}{Konrad
  Schindler}, {and} \bibinfo{person}{Vladlen Koltun}.}
  \bibinfo{year}{2022}\natexlab{}.
\newblock \showarticletitle{Towards Robust Monocular Depth Estimation: Mixing
  Datasets for Zero-Shot Cross-Dataset Transfer}.
\newblock \bibinfo{journal}{\emph{IEEE Transactions on Pattern Analysis and
  Machine Intelligence}} \bibinfo{volume}{44}, \bibinfo{number}{3}
  (\bibinfo{year}{2022}).
\newblock


\bibitem[Ravi et~al\mbox{.}(2020)]%
        {pytorch3d}
\bibfield{author}{\bibinfo{person}{Nikhila Ravi}, \bibinfo{person}{Jeremy
  Reizenstein}, \bibinfo{person}{David Novotny}, \bibinfo{person}{Taylor
  Gordon}, \bibinfo{person}{Wan-Yen Lo}, \bibinfo{person}{Justin Johnson},
  {and} \bibinfo{person}{Georgia Gkioxari}.} \bibinfo{year}{2020}\natexlab{}.
\newblock \showarticletitle{Accelerating 3D Deep Learning with PyTorch3D}.
\newblock \bibinfo{journal}{\emph{arXiv:2007.08501}} (\bibinfo{year}{2020}).
\newblock


\bibitem[Sch\"{o}nberger and Frahm(2016)]%
        {sfm1}
\bibfield{author}{\bibinfo{person}{Johannes~Lutz Sch\"{o}nberger} {and}
  \bibinfo{person}{Jan-Michael Frahm}.} \bibinfo{year}{2016}\natexlab{}.
\newblock \showarticletitle{Structure-from-Motion Revisited}. In
  \bibinfo{booktitle}{\emph{Conference on Computer Vision and Pattern
  Recognition (CVPR)}}.
\newblock


\bibitem[Sch\"{o}nberger et~al\mbox{.}(2016)]%
        {sfm2}
\bibfield{author}{\bibinfo{person}{Johannes~Lutz Sch\"{o}nberger},
  \bibinfo{person}{Enliang Zheng}, \bibinfo{person}{Marc Pollefeys}, {and}
  \bibinfo{person}{Jan-Michael Frahm}.} \bibinfo{year}{2016}\natexlab{}.
\newblock \showarticletitle{Pixelwise View Selection for Unstructured
  Multi-View Stereo}. In \bibinfo{booktitle}{\emph{European Conference on
  Computer Vision (ECCV)}}.
\newblock


\bibitem[Snavely et~al\mbox{.}(2008)]%
        {bundle_adjustment}
\bibfield{author}{\bibinfo{person}{Noah Snavely}, \bibinfo{person}{Steven~M.
  Seitz}, {and} \bibinfo{person}{Richard Szeliski}.}
  \bibinfo{year}{2008}\natexlab{}.
\newblock \showarticletitle{Modeling the World from Internet Photo
  Collections}.
\newblock \bibinfo{journal}{\emph{Int. J. Comput. Vis.}} \bibinfo{volume}{80},
  \bibinfo{number}{2} (\bibinfo{year}{2008}), \bibinfo{pages}{189--210}.
\newblock


\bibitem[Sun et~al\mbox{.}(2021)]%
        {loftr}
\bibfield{author}{\bibinfo{person}{Jiaming Sun}, \bibinfo{person}{Zehong Shen},
  \bibinfo{person}{Yuang Wang}, \bibinfo{person}{Hujun Bao}, {and}
  \bibinfo{person}{Xiaowei Zhou}.} \bibinfo{year}{2021}\natexlab{}.
\newblock \showarticletitle{{LoFTR}: Detector-Free Local Feature Matching with
  Transformers}.
\newblock \bibinfo{journal}{\emph{CVPR}} (\bibinfo{year}{2021}).
\newblock


\bibitem[Tang et~al\mbox{.}(2022)]%
        {quadtree}
\bibfield{author}{\bibinfo{person}{Shitao Tang}, \bibinfo{person}{Jiahui
  Zhang}, \bibinfo{person}{Siyu Zhu}, {and} \bibinfo{person}{Ping Tan}.}
  \bibinfo{year}{2022}\natexlab{}.
\newblock \showarticletitle{QuadTree Attention for Vision Transformers}.
\newblock \bibinfo{journal}{\emph{ICLR}} (\bibinfo{year}{2022}).
\newblock


\bibitem[Teed and Deng(2020)]%
        {raft}
\bibfield{author}{\bibinfo{person}{Zachary Teed} {and} \bibinfo{person}{Jia
  Deng}.} \bibinfo{year}{2020}\natexlab{}.
\newblock \showarticletitle{Raft: Recurrent all-pairs field transforms for
  optical flow}. In \bibinfo{booktitle}{\emph{Computer Vision--ECCV 2020: 16th
  European Conference, Glasgow, UK, August 23--28, 2020, Proceedings, Part II
  16}}. Springer, \bibinfo{pages}{402--419}.
\newblock


\bibitem[Varma et~al\mbox{.}(2023)]%
        {GNT}
\bibfield{author}{\bibinfo{person}{Mukund~T Varma}, \bibinfo{person}{Peihao
  Wang}, \bibinfo{person}{Xuxi Chen}, \bibinfo{person}{Tianlong Chen},
  \bibinfo{person}{Subhashini Venugopalan}, {and} \bibinfo{person}{Zhangyang
  Wang}.} \bibinfo{year}{2023}\natexlab{}.
\newblock \showarticletitle{Is Attention All That Ne{RF} Needs?}. In
  \bibinfo{booktitle}{\emph{The Eleventh International Conference on Learning
  Representations}}.
\newblock


\bibitem[Wang et~al\mbox{.}(2023)]%
        {sparse-nerf}
\bibfield{author}{\bibinfo{person}{Guangcong Wang}, \bibinfo{person}{Zhaoxi
  Chen}, \bibinfo{person}{Chen~Change Loy}, {and} \bibinfo{person}{Ziwei Liu}.}
  \bibinfo{year}{2023}\natexlab{}.
\newblock \showarticletitle{SparseNeRF: Distilling Depth Ranking for Few-shot
  Novel View Synthesis}.
\newblock \bibinfo{journal}{\emph{IEEE/CVF International Conference on Computer
  Vision (ICCV)}} (\bibinfo{year}{2023}).
\newblock


\bibitem[Wang et~al\mbox{.}(2021)]%
        {nerfmm}
\bibfield{author}{\bibinfo{person}{Zirui Wang}, \bibinfo{person}{Shangzhe Wu},
  \bibinfo{person}{Weidi Xie}, \bibinfo{person}{Min Chen}, {and}
  \bibinfo{person}{Victor~Adrian Prisacariu}.} \bibinfo{year}{2021}\natexlab{}.
\newblock \showarticletitle{Ne{RF}$--$: Neural Radiance Fields Without Known
  Camera Parameters}.
\newblock \bibinfo{journal}{\emph{arXiv preprint arXiv:2102.07064}}
  (\bibinfo{year}{2021}).
\newblock


\bibitem[Wu et~al\mbox{.}(2023)]%
        {scanerf}
\bibfield{author}{\bibinfo{person}{Xiuchao Wu}, \bibinfo{person}{Jiamin Xu},
  \bibinfo{person}{Xin Zhang}, \bibinfo{person}{Hujun Bao},
  \bibinfo{person}{Qixing Huang}, \bibinfo{person}{Yujun Shen},
  \bibinfo{person}{James Tompkin}, {and} \bibinfo{person}{Weiwei Xu}.}
  \bibinfo{year}{2023}\natexlab{}.
\newblock \showarticletitle{ScaNeRF: Scalable Bundle-Adjusting Neural Radiance
  Fields for Large-Scale Scene Rendering}.
\newblock \bibinfo{journal}{\emph{ACM Trans. Graph.}} \bibinfo{volume}{42},
  \bibinfo{number}{6}, Article \bibinfo{articleno}{261} (\bibinfo{date}{dec}
  \bibinfo{year}{2023}), \bibinfo{numpages}{18}~pages.
\newblock
\showISSN{0730-0301}


\bibitem[Xing et~al\mbox{.}(2022)]%
        {rgb-xy}
\bibfield{author}{\bibinfo{person}{Jiankai Xing}, \bibinfo{person}{Fujun Luan},
  \bibinfo{person}{Ling-Qi Yan}, \bibinfo{person}{Xuejun Hu},
  \bibinfo{person}{Houde Qian}, {and} \bibinfo{person}{Kun Xu}.}
  \bibinfo{year}{2022}\natexlab{}.
\newblock \showarticletitle{Differentiable Rendering using RGBXY Derivatives
  and Optimal Transport}.
\newblock \bibinfo{journal}{\emph{ACM Trans. Graph.}} \bibinfo{volume}{41},
  \bibinfo{number}{6}, Article \bibinfo{articleno}{189} (\bibinfo{date}{dec}
  \bibinfo{year}{2022}), \bibinfo{numpages}{13}~pages.
\newblock


\bibitem[Xiong et~al\mbox{.}(2023)]%
        {sparse_gs}
\bibfield{author}{\bibinfo{person}{Haolin Xiong}, \bibinfo{person}{Sairisheek
  Muttukuru}, \bibinfo{person}{Rishi Upadhyay}, \bibinfo{person}{Pradyumna
  Chari}, {and} \bibinfo{person}{Achuta Kadambi}.}
  \bibinfo{year}{2023}\natexlab{}.
\newblock \showarticletitle{SparseGS: Real-Time 360° Sparse View Synthesis
  using Gaussian Splatting}.
\newblock \bibinfo{journal}{\emph{Arxiv}} (\bibinfo{year}{2023}).
\newblock


\bibitem[Yan et~al\mbox{.}(2023)]%
        {gs-slam}
\bibfield{author}{\bibinfo{person}{Chi Yan}, \bibinfo{person}{Delin Qu},
  \bibinfo{person}{Dong Wang}, \bibinfo{person}{Dan Xu},
  \bibinfo{person}{Zhigang Wang}, \bibinfo{person}{Bin Zhao}, {and}
  \bibinfo{person}{Xuelong Li}.} \bibinfo{year}{2023}\natexlab{}.
\newblock \showarticletitle{GS-SLAM: Dense Visual SLAM with 3D Gaussian
  Splatting}.
\newblock \bibinfo{journal}{\emph{arXiv preprint arXiv:2311.11700}}
  (\bibinfo{year}{2023}).
\newblock


\bibitem[Yu et~al\mbox{.}(2021)]%
        {pixel-nerf}
\bibfield{author}{\bibinfo{person}{Alex Yu}, \bibinfo{person}{Vickie Ye},
  \bibinfo{person}{Matthew Tancik}, {and} \bibinfo{person}{Angjoo Kanazawa}.}
  \bibinfo{year}{2021}\natexlab{}.
\newblock \showarticletitle{pixelNeRF: Neural Radiance Fields From One or Few
  Images}. In \bibinfo{booktitle}{\emph{{IEEE} Conference on Computer Vision
  and Pattern Recognition, {CVPR} 2021, virtual, June 19-25, 2021}}.
  \bibinfo{publisher}{Computer Vision Foundation / {IEEE}},
  \bibinfo{pages}{4578--4587}.
\newblock


\bibitem[Yu et~al\mbox{.}(2023)]%
        {fcclip}
\bibfield{author}{\bibinfo{person}{Qihang Yu}, \bibinfo{person}{Ju He},
  \bibinfo{person}{Xueqing Deng}, \bibinfo{person}{Xiaohui Shen}, {and}
  \bibinfo{person}{Liang-Chieh Chen}.} \bibinfo{year}{2023}\natexlab{}.
\newblock \showarticletitle{Convolutions Die Hard: Open-Vocabulary Segmentation
  with Single Frozen Convolutional CLIP}. In
  \bibinfo{booktitle}{\emph{NeurIPS}}.
\newblock


\bibitem[Yu et~al\mbox{.}(2022)]%
        {MonoSDF}
\bibfield{author}{\bibinfo{person}{Zehao Yu}, \bibinfo{person}{Songyou Peng},
  \bibinfo{person}{Michael Niemeyer}, \bibinfo{person}{Torsten Sattler}, {and}
  \bibinfo{person}{Andreas Geiger}.} \bibinfo{year}{2022}\natexlab{}.
\newblock \showarticletitle{MonoSDF: Exploring Monocular Geometric Cues for
  Neural Implicit Surface Reconstruction}. In
  \bibinfo{booktitle}{\emph{Advances in Neural Information Processing Systems
  35: Annual Conference on Neural Information Processing Systems 2022, NeurIPS
  2022, New Orleans, LA, USA, November 28 - December 9, 2022}},
  \bibfield{editor}{\bibinfo{person}{Sanmi Koyejo},
  \bibinfo{person}{S.~Mohamed}, \bibinfo{person}{A.~Agarwal},
  \bibinfo{person}{Danielle Belgrave}, \bibinfo{person}{K.~Cho}, {and}
  \bibinfo{person}{A.~Oh}} (Eds.).
\newblock


\bibitem[Zhu et~al\mbox{.}(2023)]%
        {fsgs}
\bibfield{author}{\bibinfo{person}{Zehao Zhu}, \bibinfo{person}{Zhiwen Fan},
  \bibinfo{person}{Yifan Jiang}, {and} \bibinfo{person}{Zhangyang Wang}.}
  \bibinfo{year}{2023}\natexlab{}.
\newblock \bibinfo{title}{FSGS: Real-Time Few-Shot View Synthesis using
  Gaussian Splatting}.
\newblock
\newblock
\showeprint[arxiv]{2312.00451}~[cs.CV]


\bibitem[Zwicker et~al\mbox{.}(2001)]%
        {surface_splatting}
\bibfield{author}{\bibinfo{person}{Matthias Zwicker},
  \bibinfo{person}{Hanspeter Pfister}, \bibinfo{person}{Jeroen van Baar}, {and}
  \bibinfo{person}{Markus Gross}.} \bibinfo{year}{2001}\natexlab{}.
\newblock \showarticletitle{Surface Splatting}. In
  \bibinfo{booktitle}{\emph{Proceedings of the 28th Annual Conference on
  Computer Graphics and Interactive Techniques}}
  \emph{(\bibinfo{series}{SIGGRAPH '01})}. \bibinfo{publisher}{Association for
  Computing Machinery}, \bibinfo{address}{New York, NY, USA},
  \bibinfo{pages}{371–378}.
\newblock
\showISBNx{158113374X}


\bibitem[Zwicker et~al\mbox{.}(2002)]%
        {ewa}
\bibfield{author}{\bibinfo{person}{M. Zwicker}, \bibinfo{person}{H. Pfister},
  \bibinfo{person}{J. van Baar}, {and} \bibinfo{person}{M. Gross}.}
  \bibinfo{year}{2002}\natexlab{}.
\newblock \showarticletitle{EWA Splatting}.
\newblock \bibinfo{journal}{\emph{IEEE Transactions on Visualization and
  Computer Graphics}} \bibinfo{volume}{8}, \bibinfo{number}{3}
  (\bibinfo{date}{07/2002-09/2002} \bibinfo{year}{2002}),
  \bibinfo{pages}{223--238}.
\newblock


\end{thebibliography}

\newpage


\end{document}